\documentclass[journal]{IEEEtran}
\usepackage{graphics,graphicx}

\usepackage{multirow,booktabs,color,soul,threeparttable}
\definecolor{hl}{rgb}{0.75,0.75,0.75}
\sethlcolor{hl}
\usepackage{amsmath}
\usepackage{algorithm}
\usepackage[noend]{algpseudocode}
\usepackage{algpseudocode}
\usepackage{marvosym}
\usepackage{makecell}
\usepackage{bm}
\usepackage{pbox} 
\usepackage{booktabs} 
\usepackage{amssymb}
\begin{document}
%
\title{Fully Tensorized GPU-accelerated Multi-population Evolutionary Algorithm for Constrained Multiobjective Optimization Problems 
}

\author{
	Weixiong Huang,
	Rui Wang, \emph{Senior Member},
	Wenhua Li*,
	Sheng Qi,
	Tianyu Luo,
	Delong Chen,
	Tao Zhang,
	and Ling Wang, \emph{Senior Member}
	\thanks{This work was supported by the Scientific Research Project of Xiang Jiang Lab (22XJ02003), the University Fundamental Research Fund (23-ZZCX-JDZ-28), the National Science Fund for Outstanding Young Scholars (62122093), the National Natural Science Foundation of China (72071205), the National University of Defense Technology Youth Innovation Science Fund Project (ZK25-62), the science and technology innovation Program of Hunan Province (ZC23112101-10), and the Hunan Natural Science Foundation Regional Joint Project (2023JJ50490). The authors would like to thank the support by the COSTA: complex system optimization team of the College of System Engineering at NUDT.
		
	}
	\thanks{Weixiong Huang, Rui Wang, Tao Zhang, Sheng Qi, and Shichao Fan are with the College of Systems Engineering, National University of Defense Technology, Changsha, 410073, China, and the Hunan Key Laboratory of Multi-energy System Intelligent Interconnection Technology, Changsha, 410073, China. Rui Wang is also with Xiangjiang Laboratory, Changsha, 410205, China. (e-mail: huangweixiong@nudt.edu.cn; rui\_wang@nudt.edu.cn; zhangtao@nudt.edu. cn; qisheng@nudt.edu. cn; fanshichao@nudt,edu,cn.).}
	\thanks{Ling Wang is with the Department of Automation, Tsinghua University, Beijing, 100084, China (e-mail: wangling@tsinghua.edu.cn).}
	\thanks{Corresponding Author: Wenhua Li (Email:liwenhua1030@aliyun.com).}
}

\markboth{}%
{Shell \MakeLowercase{\textit{et al.}}: Bare Demo of IEEEtran.cls for IEEE Journals}
\maketitle

\begin{abstract}
Real-world constrained multiobjective optimization problems (CMOPs) are prevalent and often come with stringent time-sensitive requirements. However, most contemporary constrained multiobjective evolutionary algorithms (CMOEAs) suffer from a number of drawbacks, including complex designs, low computational efficiency, and long convergence times, which are particularly pronounced when addressing time-sensitive CMOPs. Although research on accelerating evolutionary algorithms using GPU parallelism has advanced, existing CMOEAs still face significant limitations within GPU frameworks. To overcome these challenges, this paper proposes a GPU-accelerated multi-population evolutionary algorithm, termed GMPEA. We first systematically analyze the performance bottlenecks of representative CMOEAs when implemented in a GPU environment. To address the trade-off between computational speed and solution performance, GMPEA introduces a decomposition-based multi-population approach that is fully parallelized across its entire workflow. We conducted comparative experiments on various benchmark tests and real-world applications—the Weapon Target Assignment Problems. The results demonstrate that GMPEA achieves competitive performance even without time constraints, while its computational speed significantly surpasses that of the compared algorithms. More critically, under a strict time limit, GMPEA’s performance drastically outperforms its counterparts. This work provides compelling evidence of GMPEA's superiority in solving time-sensitive CMOPs.
\end{abstract}


%
\IEEEpeerreviewmaketitle

\section{Introduction}

\IEEEPARstart{R}{eal} world constrained multiobjective optimization problems (CMOPs) are prevalent and often come with strict time-sensitive requirements, demanding high-quality solutions within extremely short periods, typically on the order of seconds or even milliseconds. Representative applications include train rescheduling in railway systems \cite{altazin2020multi}, complex weapon target assignment problems \cite{li2024comprehensive}, and multivariate pairs trading \cite{goldkamp2019evolutionary}. A CMOP can be formally defined as follows:
\begin{eqnarray}\label{e3}  
	\begin{aligned}  
		&\mathrm{Minimize} \quad F(\mathbf{x}) = (f_1(\mathbf{x}), \ldots, f_m(\mathbf{x})) \\
		&\text{s.t.} \begin{cases}  
			g_i(\mathbf{x}) \leq 0, & i = 1, \ldots, q \\
			h_i(\mathbf{x}) = 0, & i = q + 1, \ldots, l,  
		\end{cases}  
	\end{aligned}  
\end{eqnarray}  
where $\mathbf{x}$ is a decision vector with $d$ dimensions, and $F(\mathbf{x})$ represents the set of $m$ objective functions to be minimized. The functions $g(\mathbf{x})$ and $h(\mathbf{x})$ represent the $q$ inequality constraints and the $l-q$ equality constraints, respectively. 

Over the past three decades, evolutionary algorithms (EAs) have made significant progress in solving multiobjective optimization problems (MOPs). Mainstream multiobjective evolutionary algorithms (MOEAs), such as dominance-based (e.g., NSGA-II \cite{deb2002fast}, SPEA2 \cite{zitzler2001spea2}), decomposition-based (e.g., MOEA/D \cite{zhang2007moea}, MOEA/D-AAWN \cite{zhao2022decomposition}), and indicator-based (e.g., HypE \cite{zitzler2004indicator}, IBEA \cite{bader2011hype}) algorithms, face inherent challenges when directly applied to CMOPs due to their general lack of effective constraint-handling mechanisms. Consequently, a large body of excellent constraint-handling techniques has emerged over the last decade \cite{fonseca2002multiobjective, takahama2010efficient, wang2012combining, runarsson2000stochastic, jan2013study}. These fundamental techniques can be skillfully integrated into various MOEAs to balance objective optimization and constraint satisfaction, leading to the development of different types of constrained multiobjective evolutionary algorithms (CMOEAs), which can be broadly classified into three categories: (1) feasibility-first CMOEAs; (2) constraint relaxation-based CMOEAs; and (3) multi-stage, multi-population, or multi-task CMOEAs.

Although many CMOEAs have demonstrated good performance in solving a variety of CMOPs, their efficacy remains limited by the underlying computational capacity. Most CMOEAs rely heavily on CPUs, and their inherent serial computation limits overall efficiency, particularly when tackling CMOPs that require high-speed solutions. To address this challenge, two main strategies are typically adopted in the field: (1) reducing the algorithm's scale, such as decreasing the number of iterations or the population size to achieve faster solutions, which often leads to a significant degradation in performance; or (2) employing learning-based methods to quickly generate approximate solutions via a trained model \cite{song2023balancing,  wu2024multi}. However, learning-based approaches face considerable challenges in terms of generalization, reliability, and stability, making it difficult to guarantee their effectiveness in unknown or dynamic environments.

Recently, many researchers have sought to harness the GPU’s massive parallelism to accelerate evolutionary algorithms. For example, Huang et al. \cite{huang2024evox} proposed EvoX, a GPU‑accelerated platform for evolutionary optimization. By leveraging the high concurrency enabled by tensorization, this platform opens a new avenue for rapidly solving MOPs. However, EvoX has not yet been extended to CMOPs or CMOEAs, and GPU‑accelerated CMOEAs remain underexplored. Therefore, we systematically ported and reproduced several representative CMOEAs and CMOPs on the EvoX platform and conducted an in‑depth analysis of their performance in a GPU‑accelerated setting. Our findings reveal that most CMOEAs fail to realize the expected speedups on GPUs. The primary reason is their reliance on complex environmental‑selection strategies with inherent serial dependencies and tight computational coupling, which prevents true end‑to‑end parallelism. For instance, many CMOEAs adopt the SPEA2 truncation strategy for survivor selection; this procedure is inherently sequential, as removing each individual requires on‑the‑fly recomputation and each step depends on the outcome of the previous one. Moreover, many CMOEAs follow a “small‑population, many‑iterations” paradigm that is ill‑suited to time‑sensitive CMOPs. Iteration‑driven adaptive mechanisms (e.g., constraint relaxation and multi‑stage schemes) can also become ineffective under the GPU‑friendly “large‑population, few‑iterations” regime.

Building on the above considerations, this paper proposes a decomposition-based GPU-accelerated multi-population evolutionary algorithm (GMPEA). The specific contributions of this paper are as follows:

  \begin{enumerate}[]
  	\item Platform Extension and CMOEA Performance Analysis: We extend the EvoX platform to support constraint handling by systematically porting and reproducing a range of representative CMOEAs and CMOPs. Based on this, we conduct an in-depth analysis of the performance bottlenecks and algorithmic limitations that mainstream CMOEAs face in a GPU parallel environment.
  	
  	\item Tensorized Algorithm Design for Efficiency and Performance: To address the common trade-off between solution quality and computational efficiency in existing CMOEAs when solving time-sensitive problems, we propose a fully tensorized, decomposition-based dual-population algorithm. This algorithm achieves fully tensorization, significantly improving solution quality while ensuring rapid convergence. Its effectiveness is validated through extensive experiments on a variety of standard benchmark problems.
  	
  	\item Differentiated Neighborhood Strategy for Parallel Architectures: In decomposition-based algorithms, the choice of neighborhood size has always been a key challenge: a large neighborhood weakens local search, while a small neighborhood can impede convergence. Furthermore, in the “few iterations” GPU optimization paradigm, adaptive neighborhood sizing strategies are often ineffective. Our proposed decomposition-based dual-population algorithm cleverly resolves this issue through a differentiated neighborhood strategy: the unconstrained population uses a large neighborhood for rapid convergence, while the constraint-handling population employs a small neighborhood to enhance local diversity.

  \end{enumerate}

The rest of this paper is organized as follows. Section \ref{s2} reviews some background knowledge and related work. Section \ref{s3} details the design and implementation of the proposed GMPEA. Sections \ref{s4} and \ref{s5}, respectively, outline the experimental setup and analyze the results. Finally, Section \ref{s6} concludes this paper and discusses future directions.

\section{Background}
\label{s2}
\subsection{Basic Definitions} 
\subsubsection{Constraint Violation}
In CMOPs, evaluating a solution extends beyond its objective values to include its constraint satisfaction. The violation of a solution $\mathbf{x}$ on the $j$th constraint, denoted as $C_j(\mathbf{x})$, is defined as:
\begin{equation}
	\label{e3}
	\textit{C}_\textit{j}(\mathbf{x})=\left\{\begin{array}{l}
		\max \left(0, g_j(\mathbf{x})\right), j=1, \ldots, p \\
		\max \left(0,\left|h_j(\mathbf{x})\right|-\delta\right), j=p+1, \ldots, p+q
	\end{array}\right.
\end{equation}	
Here, the variable $\delta$ is a small tolerance (typically set to $10^{-6}$) used to relax the strictness of equality constraints, effectively converting them into inequality constraints. The total constraint violation, $CV(\mathbf{x})$, is the sum of all individual constraint violations:
\begin{equation}
	\label{e4}
	\textit{CV}(\mathbf{x})=\sum_{j=1}^{p+q} \textit{C}_\textit{j}(\mathbf{x}),
\end{equation}
a solution $\mathbf{x}$ is considered feasible if its total constraint violation is $CV(\mathbf{x})=0$, and infeasible otherwise.

\subsubsection{Feasible Region}
A solution $\mathbf{x}$ is considered feasible if it satisfies all constraints, i.e., $CV(\mathbf{x})=0$. The collection of all such feasible solutions forms the feasible region, denoted as:
\begin{equation}
	\label{e5}
	S=\{\mathbf{x}\mid \textit{CV}(\mathbf{x})=0, \mathbf{x} \in \Omega\}.
\end{equation}

\subsubsection{Pareto Dominance}
Given two solutions, $\mathbf{x}$ and $\mathbf{y}$, we formally define the Pareto dominance relationship, denoted as $\mathbf{x} \prec \mathbf{y}$, as follows:
\begin{equation}
	\label{e6}
	\begin{cases}
		\forall i\in\{1,2,\ldots,M\}:f_i(\mathbf{x})\leq f_i(\mathbf{y}) \\
		\exists j\in\{1,2,\ldots,M\}:f_j(\mathbf{x})<f_j(\mathbf{y})
	\end{cases}.
\end{equation}	
This definition establishes that $\mathbf{x}$ dominates $\mathbf{y}$ if it is not inferior to $\mathbf{y}$ on any objective and is strictly better on at least one objective.

\subsubsection{Constraint Dominance Principle (CDP) \cite{deb2002fast}}
For two solutions, $\mathbf{x}$ and $\mathbf{y}$, $\mathbf{x}$ is said to dominate $\mathbf{y}$ based on the CDP if and only if any of the following conditions are satisfied:
\begin{equation}
	\label{e7}
	\begin{cases}
		CV(\mathbf{x})=0 \text{ and } CV(\mathbf{y})>0 \\
		CV(\mathbf{x})=0, CV(\mathbf{y})=0, \text{ and } \mathbf{x} \prec \mathbf{y} \\
		CV(\mathbf{x})>0, CV(\mathbf{y})>0, \text{ and } CV(\mathbf{x})<CV(\mathbf{y})
	\end{cases}.
\end{equation}

\subsubsection{Aggregation Function}

Decomposition-based MOEAs often utilize aggregation functions, such as the weighted sum, Tchebycheff, and penalty-based boundary intersection (PBI) functions to update individuals within a neighborhood. This paper primarily employs the PBI function, which is defined as follows:

\begin{equation}
	\label{e8}
	f^{\textit{pbi}}(F,W,z)=d_1+\theta\cdot d_2,
\end{equation}
where $d_1=\frac{\left\|(F-z)^\top\cdot W \right\|}{\|W\|}$, $d_2=\|F-(z+d_1\cdot W)\|$. Here, $\|\cdot\|$ denotes the Euclidean norm, $\theta$ is a predefined penalty parameter (typically set to 5), $W$ is the reference vector, and $z$ is the ideal point.

For unconstrained MOPs, neighborhood individuals are updated by a direct comparison of their aggregation function values. For CMOPs, to ensure the feasibility of the final solutions, a feasibility priority rule (FPR) \cite{jain2013evolutionary} is used for comparison. For two solutions, $\mathbf{x}$ and $\mathbf{y}$, the rule states that $\mathbf{x}$ is considered better than $\mathbf{y}$ if any of the following conditions are met:

\begin{equation}
	\begin{cases}
		f^{pbi}(\mathbf{x}) < f^{pbi}(\mathbf{y}), & \text{if} \quad CV(\mathbf{x}) = CV(\mathbf{y}) \\
		CV(\mathbf{x}) < CV(\mathbf{y}), & \text{if} \quad CV(\mathbf{x}) \neq CV(\mathbf{y})
	\end{cases}
\end{equation}

\subsubsection{Tensorization Operations}
A tensor is essentially a multidimensional array, and its inherent parallelism makes it well-suited for computation on GPUs. By transforming certain processes, operations, and data structures within algorithms into tensor representations, we can eliminate explicit loops (provided that the operations within the loops are independent) and conditional branches, which promotes efficient computation and optimizes hardware utilization \cite{liang2025bridging}. 

Common tensor operations include tensor multiplication, expressed as \( \boldsymbol{A} \cdot \boldsymbol{B} \), and the Hadamard product, denoted as \( \boldsymbol{A} \odot \boldsymbol{B} \), which performs element-wise multiplication at specified positions. The Heaviside step function, represented as \( H(\boldsymbol{A}) \), returns a tensor that is isomorphic to \( \boldsymbol{A} \); it yields 1 if the corresponding elements in \( \boldsymbol{A} \) are greater than or equal to 0, and 0 otherwise.  Additionally, tensor operations frequently utilize advanced functions such as sort, argsort, min, max, and argmin. The vmap (vectorized map) function is designed to efficiently handle batch data, particularly in the context of neural network training; it achieves this by mapping a specific operation to multiple instances of input data, thereby eliminating the need for explicit loops and optimizing performance on hardware that supports vectorization. Mathematically, `vmap` automates batch processing of functions. If we let \( f: \mathbb{R}^n \rightarrow \mathbb{R}^m \) be a function, then `vmap` can be applied to a batch of input data points \( \boldsymbol{X} \in \mathbb{R}^{k \times n} \), yielding an output batch of data \( \boldsymbol{Y} \in \mathbb{R}^{k \times m} \): 

\begin{equation}
\boldsymbol{Y} = \mathbf{vmap}(f)(\boldsymbol{X}) = \begin{bmatrix}
	f(\boldsymbol{x}_1) \\
	f(\boldsymbol{x}_2) \\
	\vdots \\
	f(\boldsymbol{x}_k)
\end{bmatrix},
\end{equation}
where \( \boldsymbol{x}_i \) is the \( i \)-th input sample in the batch \( \boldsymbol{X} \).

\subsection{Classification of CMOEAs}

In the research conducted over the past few decades, existing CMOEAs can primarily be categorized into three main types:

Feasibility prioritized CMOEAs: These algorithms focus primarily on generating feasible solutions, often granting higher priority to feasible individuals during selection, mating, or offspring updates. However, excessive emphasis on feasibility can result in a loss of population diversity, leading to premature convergence in local feasible region and restricting exploration of the global optimal front. The constrained dominance principle (CDP), as established by Deb et al. \cite{deb2002fast}, asserts that any feasible solution is superior to any infeasible one. 
Coello et al. \cite{coello1999moses} proposed a method that eliminates infeasible solutions from the population; however, this approach may remove valuable search directions when the entire population consists of infeasible solutions. Furthermore, researchers have explored more sophisticated approaches. Ma et al. \cite{ma2019new} developed a novel fitness measure that incorporates both CDP and Pareto dominance rankings. Yu et al. \cite{yu2021dynamic} proposed a dynamic, preference-based selection method to balance constraint satisfaction and objective optimization. 

Constraint relaxation Based CMOEAs: To alleviate the high pressure for feasibility selection, these algorithms intentionally relax constraints during early evolutionary phases or at specific stages, allowing for a degree of constraint violation. The primary aim is to expand the search space and leverage high-quality infeasible solutions to guide the population toward feasibility in later stages. This strategy helps to escape local optima and enhances global exploration capabilities.Woldesenbet et al. \cite{woldesenbet2009constraint} pioneered a method that combines an adaptive penalty function with a specific distance metric. Chih-Hao et al. \cite{lin2013rough} introduced a penalty function integrated with rough set theory, enabling adaptive adjustment of the penalty coefficient based on constraint violations. PPS \cite{fan2019push} adjusts constraints through a “push-pull” process by modifying the $\varepsilon$ value, alternating between tightening and relaxing constraint boundaries. Fan et al. \cite{fan2017improved} proposed a dynamic $\varepsilon$ approach, adjusting the value according to the proportion of feasible solutions within the population, particularly effective for CMOPs with extensive infeasible regions. Zhu et al. \cite{zhu2020constrained} developed a variant of MOEA/D, named MOEA/D-DAE, designed to prevent premature convergence by implementing a “detect-and-escape” strategy that expands constraint boundaries when the population becomes trapped, facilitating renewed exploration.

Multi-Population, Multi-Stage, or Multi-Task CMOEAs: These algorithms enhance the adaptability to complex CMOPs by either coordinating different populations in parallel or by dividing the evolutionary process into sequential stages. This allows various search strategies or constraint-handling mechanisms to run concurrently or alternately, providing a more flexible and robust approach to solving difficult problems. 
Several multi-stage algorithms have been developed to manage this complexity. Liu et al. \cite{liu2019handling} introduced a two-stage algorithm that converts a CMOP into a single-objective problem to accelerate convergence.  Ma et al. \cite{ma2021multi} proposed a multi-stage method that manages constraints incrementally across different phases of the evolutionary process. CMOEA-MS \cite{tian2021balancing} utilizes varying fitness strategies, initially treating objectives equally before prioritizing constraints in later phases. Sun et al. \cite{sun2022multistage} developed the C3M algorithm, which focuses on handling interrelated constraints to improve exploration and solution quality without applying strict feasibility constraints early on.
Other works have explored multi-population strategies. Li et al. \cite{li2018two} presented C-TAEA, which uses two separate archives to optimize feasible solutions and promote broader exploration simultaneously. Similarly, Tian et al. \cite{tian2020coevolutionary} introduced CCMO, which employs a dual-population structure with minimal collaboration: one population focuses on the original CMOP while the other handles an unconstrained version. Zou et al. \cite{zou2023multipopulation} proposed MCCMO, where each population is dedicated to handling distinct constraints, with collaboration mechanisms to share information.
Finally, multi-task frameworks offer another sophisticated approach to handling CMOPs. Qiao et al. \cite{qiao2022evolutionary} developed EMCMO, which transforms the optimization into two related tasks with knowledge transfer between them. IMTCMO \cite{qiao2023evolutionary} enhance algorithm performance by using an auxiliary task to incorporate infeasible solutions and improve local and global search strategies. Ming et al. \cite{ming2023adaptive} introduced a multitasking framework that leverages reinforcement learning to dynamically select auxiliary tasks, resulting in two new algorithms based on Q-Learning and Deep Q-Learning.

\subsection{GPU-Accelerated MOEAs}
In recent years, significant advancements in GPU hardware and parallel computing have prompted many researchers to utilize GPUs for accelerating complex calculations on large-scale datasets. While this trend has gained considerable traction, research focusing on GPU-accelerated CMOEAs is still in its early stages. Nevertheless, steady progress has been made in the realm of unconstrained MOEAs .

Early efforts primarily targeted the acceleration of key components within existing algorithms. Notably, Wong \cite{wong2009parallel} pioneered the integration of GPU-accelerated non-dominated sorting into the NSGA-II framework. This foundational work was later expanded by Sharma et al. \cite{sharma2010gpgpu}, who introduced GASREA, a variant that offloads the sorting of an external non-dominated archive to the GPU. Aguilar-Rivera \cite{aguilar2020gpu} further advanced this field with a fully vectorized version of NSGA-II, employing randomized sorting and grid-based crowding techniques.

Beyond NSGA-II, GPU acceleration has been effectively applied to various other evolutionary algorithms. For example, Souza et al. \cite{de2014gpu} successfully implemented a GPU-accelerated version of the MOEA/D-ACO algorithm, while Hussain et al. \cite{hussain2020gpu} showcased a rapid CUDA-based implementation of Multi-Objective Particle Swarm Optimization (MOPSO). Wang et al. \cite{wang2022tensor} enhanced Particle Swarm Optimization by incorporating tensor models to tackle multiobjective and dynamic optimization challenges, and Lopez et al. \cite{lopez2015gpu} improved the performance of SMS-EMOA by accelerating the computationally intensive hypervolume contribution calculations using GPU resources.

More recently, there has been a shift from algorithm-specific implementations to more flexible, general-purpose frameworks. Huang et al. \cite{huang2024evox} developed the EvoX platform based on Google's JAX, establishing a GPU-accelerated framework for general evolutionary computation. Building upon EvoX, Li et al. \cite{li2025gpu} proposed TensorNSGA-III, a fully tensorized implementation of NSGA-III that retains the original algorithm's selection and mutation mechanisms. Liang et al. \cite{liang2025bridging} introduced a tensorization method that restructured traditional MOEAs for GPU acceleration, achieving significant speedups when addressing large-scale MOPs. Furthermore, Liang et al. \cite{liang2024gpu} presented the first fully tensorized implementation of a reference vector-guided MOEA based on RVEA.

Building on these foundational advancements, our research harnesses the parallel computing capabilities and scalability of the EvoX platform to investigate the potential of GPU acceleration for CMOEAs, culminating in the work presented in this paper.

\subsection{Motivation}

The key to fully exploiting the parallel computing capabilities of GPUs lies in tensorization \cite{wang2024tensorized}, which refers to the representation of data and operations using high-dimensional matrices. By expressing populations and related operations as tensors, the performance of MOEAs can be significantly accelerated. While some preliminary studies \cite{liang2024gpu, wang2021tensor} have shown that tensorization can substantially improve computational speed and accommodate larger populations, fully tensorized MOEAs remain underdeveloped, particularly in the context of CMOEAs.

To ensure efficient acceleration, MOEAs must implement complete tensorization across all stages, including initialization, matching selection, offspring updating, and environmental selection. However, the current design of environmental selection strategies in CMOEAs tends to be complex, creating significant challenges for achieving thorough tensorization.

\begin{table*}[htbp]
	\centering
	\caption{Inherent Limitations of Representative CMOEAs for GPU Acceleration}
	\setlength{\tabcolsep}{3pt}
	\begin{tabular}{c|p{10em}|p{15em}|p{30em}}
		\toprule
		\textbf{Algorithms} & \textbf{CMOEAs Type} & \textbf{Environment Selection Operation} & \multicolumn{1}{c}{\textbf{Limitations in GPU Environment}} \\
		\midrule
		NSGA-II-CDP & Feasibility prioritized & \pbox{18em}{1. Non-dominated sorting\\2. Crowding distance calculation} & In the early stages of solving CMOPs, algorithms often encounter a large number of infeasible solutions. This causes the number of non-dominated ranks to increase sharply, which in turn significantly diminishes the acceleration benefits of tensorization. \\
		\midrule
		PPS & \pbox{18em}{Constraint relaxation \\ Multi-Stage} & \pbox{18em}{1. Aggregation function calculation\\2. Neighborhood update\\3. Crowding distance calculation} & With fewer iterations, the constraint relaxation strategy may have negative effects. And the environment selection strategy for the archive set in PPS uses the non-dominated sorting method based on CDP. \\
		\midrule
		CMOEA-MS & Multi-Stage & \pbox{18em}{1. Fitness calculation\\2. Truncation operation} & Iterative dependencies and serial stage transitions are not compatible with the “large population, few iterations” GPU paradigm. Additionally, truncation operations are difficult to tensorize. \\
		\midrule
		CCMO & Multi-Population & \pbox{18em}{1. Fitness calculation\\2. Truncation operation} & The serial dependency of truncation operations makes them difficult to tensorize, which results in a slow overall algorithm. If multiple populations cannot be computed in parallel, the computational burden will increase significantly with population size. \\
		\midrule
		EMCMO & \pbox{18em}{Multi-Task \\ Multi-Stage} & \pbox{18em}{1. Fitness calculation\\2. Truncation operation} & The design is relatively complex, and truncation operations are difficult to tensorize. There is also some computational coupling between individual migrations of multiple tasks. \\
		\bottomrule
	\end{tabular}
	\label{t1}
\end{table*}

\begin{figure}[!htbp]
	\centering
	\includegraphics[width=8.8cm]{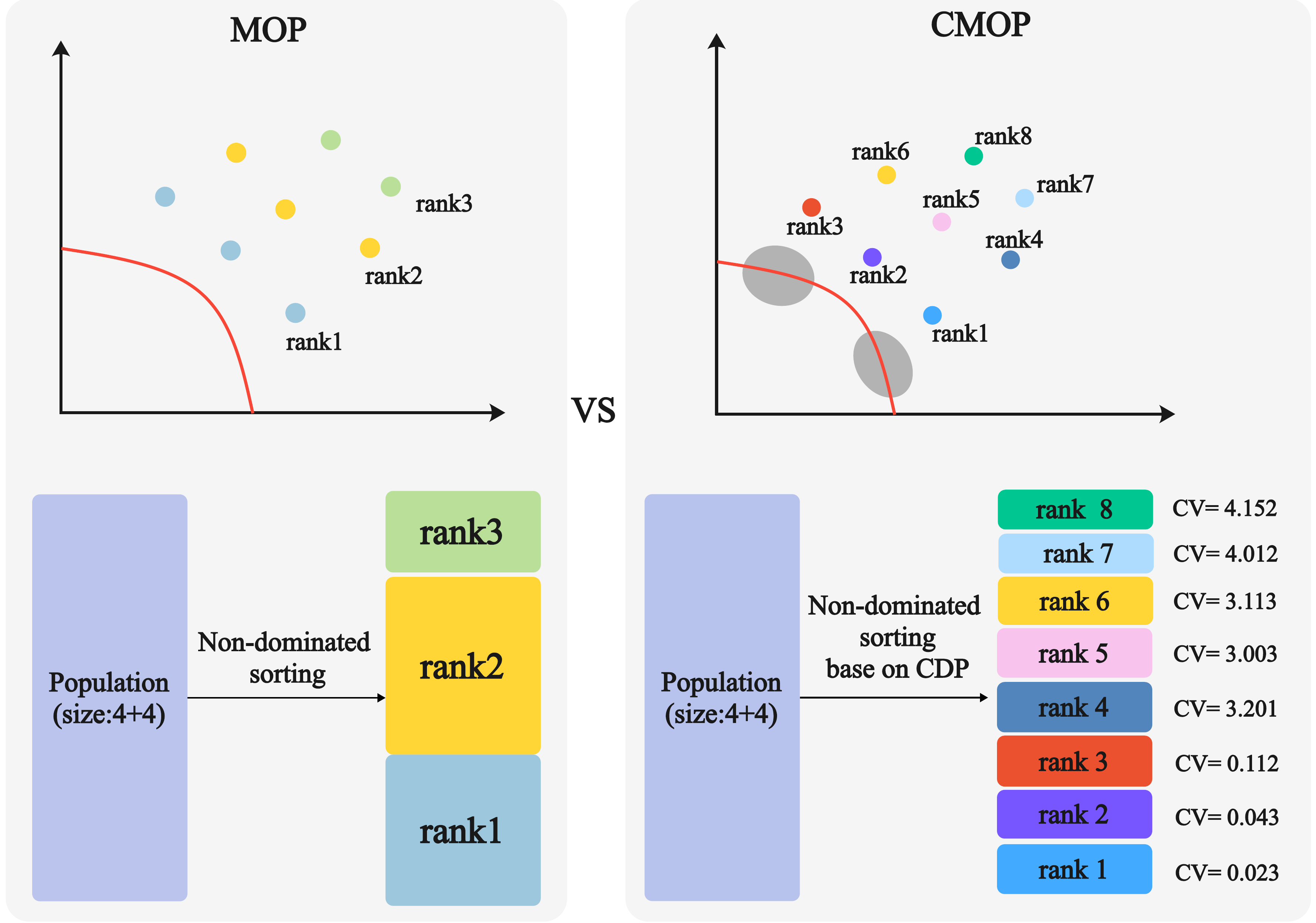}
	\centering
	
	\caption{An illustrative example highlighting the potential difference in the number of ranks after non-dominated sorting for MOPs and CMOPs. The right figure demonstrates how NSGA-II-CDP can lead to a significant increase in ranks for CMOPs, especially when many infeasible solutions exist early on, thereby diminishing the effectiveness of parallel acceleration.}
	\label{f1}
\end{figure}

To underscore the challenges of GPU acceleration for CMOEAs, we analyzed the inherent limitations of several representative algorithms when deployed in a parallel environment. As Table 1 shows, these algorithms possess design characteristics that hinder their performance on GPUs. Non-dominated sorting based algorithms (like NSGA-II-CDP \cite{deb2002fast} and ToR  \cite{ma2019new}) often face a large number of infeasible solutions early in the CMOP solving process, causing the number of non-dominated ranks to increase dramatically and pushing the algorithm's time complexity toward its worst-case scenario. The example in Fig. \ref{f1} highlights the computational differences between solving MOPs and CMOPs, showing a sharp increase in the initial computational load for CMOPs. GPU acceleration for non-dominated sorting relies on tensorization to parallelize individuals within the same rank, but when ranks grow to contain only one individual each, the acceleration benefits are significantly diminished.

Additionally, algorithms using multi-stage architectures or constraint relaxation techniques (such as CMOEA-MS \cite{tian2021balancing} and PPS \cite{fan2019push}) struggle to optimize effectively in GPU parallel environments, particularly under the “large population, few iterations” paradigm. Their iterative dependencies and serial stage transitions conflict with GPU's inherent parallelism. Many CMOEAs also employ truncation strategies derived from SPEA2 \cite{zitzler2001spea2} to maintain population diversity (e.g., CMOEA-MS \cite{tian2021balancing}, CCMO \cite{tian2020coevolutionary}, EMCMO \cite{qiao2022evolutionary}). This procedure inherently functions as a serial loop where each individual removal necessitates real-time calculations dependent on previous outcomes, creating substantial computational coupling that makes parallelization challenging.

While multi-population or multi-task structured algorithms are theoretically well-suited for parallel implementation, their computational speed remains bottlenecked by environmental selection efficiency, constraint handling techniques, and inter-population coupling. These representative CMOEAs have inherent flaws when deployed in GPU parallel environments simply transferring them to GPUs with only partial tensorization may not yield anticipated acceleration, as overall performance becomes limited by the least effective component.

To overcome these limitations and fully utilize GPU parallel advantages for solving time-sensitive CMOPs, this study proposes a novel GPU-accelerated Multi-Population decomposition-based Evolutionary Algorithm. We emphasize fully tensorized structure and operations throughout the entire process. We chose the decomposition-based structure because its design is relatively simple and subproblems have independence, facilitating decoupling and restructuring within a GPU framework. The tensorized MOEA/D proposed by Liang et al. \cite{liang2025bridging} also demonstrated its effectiveness in solving large-scale MOPs.

\section{The Proposed Algorithm}
\label{s3}
This section details the proposed GMPEA , focusing on its core components: the multi-population parallel mechanism and the tensorized implementation of the environmental selection operation. The remaining general components common to multiobjective evolutionary algorithms, such as initialization, function evaluation, mating selection, and crossover and mutation, are not elaborated here, as their efficient tensorized implementations are already available in the EvoX framework. For detailed information on these standard operations, please refer to references \cite{liang2024gpu, liang2025bridging}.
\begin{figure*}[!htbp]
	\centering
	\includegraphics[width=18.3cm]{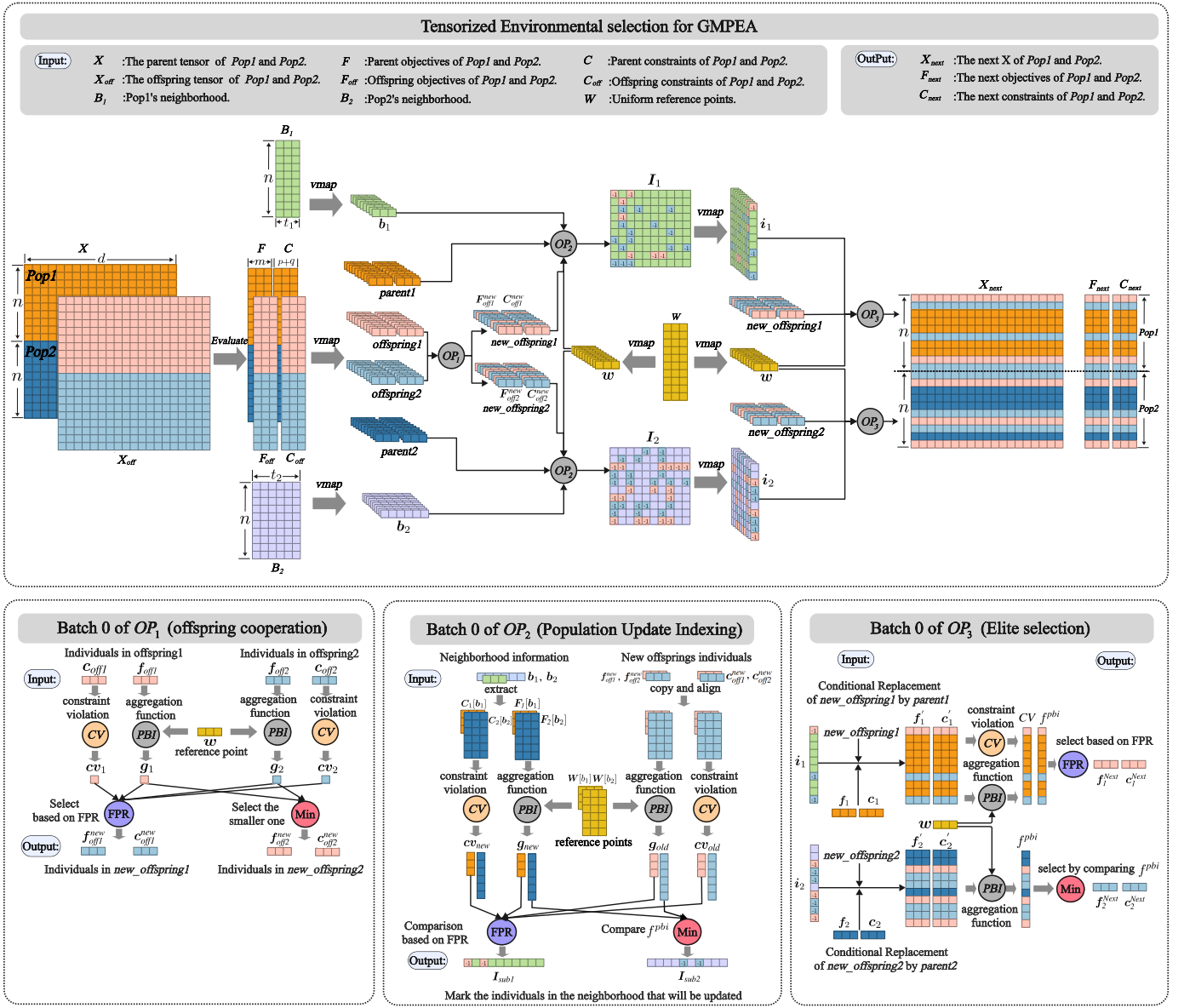}
	\centering
	
	\caption{An overview of the tensorized environmental selection process in GMPEA. The top diagram illustrates the overall tensor data flow of the environmental selection. The bottom row shows conceptual diagrams for a single batch operation within each of the three key modules: $OP_1$(Offspring Cooperation), $OP_2$(Population Update Indexing), and $OP_3$(Elite Selection).}
	\label{f2}
\end{figure*}

\subsection{Framework}

GMPEA introduces two distinct, yet cooperative, populations, a design inspired by classic dual-population algorithms \cite{tian2020coevolutionary}.  $Pop_1$ prioritizes constraint handling, with a focus on local exploration and convergence within the feasible region to enhance the algorithm's ability to handle constraints. In contrast, $Pop_2$ disregards constraints and focuses on solving the unconstrained version of the problem, aiming to promote global exploration and maintain diversity. During the environmental selection phase, these two populations exchange beneficial information to achieve collaborative evolution, striking a better balance between constraint satisfaction and objective optimization.

Algorithm \ref{a1} presents the pseudocode for the overall framework of GMPEA. All populations are represented as structured tensors. For example, $Pop_1\{\boldsymbol{X}_1, \boldsymbol{F}_1, \boldsymbol{C}_1\}$ represents the first population, $Pop_1$, with its decision variable tensor $\boldsymbol{X}_1$, objective tensor $\boldsymbol{F}_1$, and constraint tensor $\boldsymbol{C}_1$. The first line depicts the initialization procedure for the two populations, ensuring their initial distribution and size. The second line performs parallel evaluations of the two populations. The third line initializes the neighborhood structures. The iterative optimization begins in line six, where each population generates its offspring. Immediately following this, line seven creates two offspring populations for parallel function evaluation. Finally, line eight executes environmental selection based on the distinct roles of the two populations, adopting corresponding strategies. 

\subsection{Tensorized Environmental Selection}﻿

Fig. \ref{f2} visually represents the core environmental selection process of GMPEA, highlighting how the algorithm achieves efficient population updates through tensorized operations after parallel evaluation of both parent and offspring populations. The essence of this mechanism lies in its differentiated neighborhood information processing and cooperative multi-population update strategy.

In GMPEA, parent and offspring individuals from both $Pop_1$ and $Pop_2$ are structured as unified tensors to maximize the utilization of GPU parallel computing. $Pop_1$ is designed for enhanced local search and constraint consideration, whereas $Pop_2$ disregards constraints to promote global convergence. After parallel function evaluation, the objective and constraint violation values for all parent and offspring individuals are extracted and organized into separate tensors. To enable a differentiated search, GMPEA employs different neighborhood sizes for the two populations: $Pop_1$ utilizes a small neighborhood, $B_1$, to enhance local diversity, while $Pop_1$ employs a large neighborhood, $B_2$, to accelerate global convergence. This strategy addresses the trade-off between convergence and diversity that is often faced by single-population, decomposition-based algorithms.

The environmental selection in GMPEA is achieved through a series of carefully designed tensorized operations. These operations are independent at the individual level and are applied in a batch to the entire tensor via a vmap (vectorized map) operation. This approach scales up independent, individual-level operations to efficient, tensor-level processing, thereby avoiding explicit serial loops. To further enhance parallel computation, all traditional conditional branches are transformed into tensorization mask operations. For example, a conditional branch with condition tensor $\boldsymbol{C}$ and outcome tensors $\boldsymbol{A}$ and $\boldsymbol{B}$, which produces output tensors $\boldsymbol{Y}$, can be expressed as:
\begin{equation}
Y= H(\boldsymbol{C})\odot \boldsymbol{A}+(1-H(\boldsymbol{C}))\odot \boldsymbol{B},
\end{equation}
here, H() is the Heaviside step function.

Algorithm \ref{a2} provides further details on the pseudocode for GMPEA's environmental selection process.  The process is composed of the following three main operations.

\subsubsection{\textbf{Offspring Cooperation} (OP\textsubscript{1})}
This operation is designed to select the most promising offspring individuals for each subproblem in both populations. It parallelizes the individual-level comparison logic using a vmap operation, expressed as:
\begin{equation}
	Off^\text{new}_1, Off^\text{new}_2 \gets \mathbf{vmap}(OP_1)(Off_1, Off_2).
\end{equation}
For each subproblem, it compares the offspring from both populations with their respective parent individuals. The selection logic for $Pop_2$, which disregards constraints, is straightforward: it selects the offspring with a lower aggregated function value to accelerate convergence. For $Pop_1$, which considers constraints, the selection logic is more complex. It employs the FPR to consider both constraint violation and the aggregated function value. $Pop_1$ first prioritizes feasible solutions; if both individuals are either feasible or infeasible, it then compares their aggregated function values. Finally, $OP_1$ outputs two new, pre-filtered offspring population tensors, $Off^\text{new}_1$ and $Off^\text{new}_2$.

\subsubsection{\textbf{Population Update Indexing} (OP\textsubscript{2})}
The core objective of this operation is to determine which parent individuals will be replaced by new offspring. It generates update indices based on two different sets of comparison rules between the parent and offspring populations. This process ultimately creates update index tensors for both populations, where each row corresponds to a subproblem, expressed as:
\begin{equation}
	\boldsymbol{I}_1, \boldsymbol{I}_2 \gets \mathbf{vmap}(OP_2)(\boldsymbol{B}_{1}, \boldsymbol{B}_{1}, Off^\text{new}_1, Off^\text{new}_2).
\end{equation}
The indices in the tensor indicate which parent individuals should be updated, with a specific value (e.g., -1) representing the update positions. This provides an efficient and parallelizable data structure to guide subsequent population updates. It is important to note that from a column perspective, a parent individual may be identified as the best candidate for replacement by multiple offspring. Therefore, the results of this step may contain conflicts, which require further screening by $OP_3$. $Pop_1$ and $Pop_2$ use their respective comparison rules in this step: $Pop_1$ relies on the FPR, while $Pop_2$ depends solely on the aggregated function values.

\subsubsection{\textbf{Elite Selection and Final Population Update} (OP\textsubscript{3})}
This operation is designed to resolve the potential conflicts from $OP_2$ and complete the final population update, expressed as:
\begin{equation}
	Pop^\text{next}_{1},Pop^\text{next}_{2}\gets\mathbf{vmap}(OP_3)(\boldsymbol{I}^{\top}_1, \boldsymbol{I}^{\top}_2,Pop_{1},Pop_{2}).
\end{equation}
For each subproblem, when its parent individual might be updated by multiple offspring, $OP_3$ selects the best elite offspring from these candidates to replace the parent based on their respective comparison rules. $Pop_1$ continues to use the FPR to select the best offspring, ensuring a balance between constraint satisfaction and objective optimization in the new population. $Pop_2$, in contrast, continues to use the aggregated function value as its sole criterion to retain the offspring that best promotes convergence. Through this fully parallel elite selection process, two new populations, Next $Pop_1$ and Next $Pop_2$, are efficiently constructed.

This completely tensorized, multi-population parallel approach, with its differentiated neighborhood strategy and cooperative environmental selection, ensures GMPEA's ability to overcome the performance bottlenecks of existing CMOEAs in GPU environments. This gives GMPEA a significant performance advantage when solving high-timeliness CMOPs.

\subsection{Discussion}
While multi-population algorithms are a common paradigm in evolutionary computation, most existing ones are designed for CPU-dependent computing environments. Their designs are often complex, especially when multi-population structures are coupled with multi-stage strategies, making the algorithm difficult to decouple and refactor for a fully tensorized implementation. This complexity creates a degree of coupling between populations (especially in information exchange), hindering parallelization. More importantly, as we discussed in the motivation section, a key bottleneck of existing CMOEAs is the inherent limitation in their environmental selection phase, which prevents efficient acceleration on GPUs. Many algorithms rely on typical serial computation strategies (such as the pruning strategy from SPEA2 [78]) to maintain population distribution. These are sequential deletion processes where each operation depends on the result of the previous one, making them impossible to parallelize. Furthermore, when solving CMOPs, non-domination-based algorithms (like NSGA-II) face an explosive increase in non-dominated layers due to a large number of initial infeasible solutions. This causes the time complexity of the sorting algorithm to easily reach its worst-case scenario, leading to a sharp increase in computation. These factors collectively prevent existing algorithms from achieving significant acceleration on GPUs, which is a critical drawback for solving high-timeliness problems.

The design of GMPEA is based on a deep understanding of these challenges. By comprehensively considering both the algorithmic paradigm and the underlying implementation, we have fundamentally addressed the inherent limitations of existing CMOEAs in GPU environments. The core idea is to design a highly decoupled, parallel-friendly algorithm structure and abandon traditional serial environmental selection strategies. We adopt a decomposition-based, multi-population structure, breaking down complex CMOP tasks into relatively independent subproblems. We then designed a fully tensorized environmental selection process that is well-suited for multiple populations. This process transforms all comparison, screening, and update operations into large-scale tensor computations, utilizing functions like vmap to enable independent, parallel processing for each subproblem. This design not only avoids complex serial dependencies but also resolves the neighborhood-setting dilemma through a differentiated neighborhood strategy (large neighborhoods for convergence, small ones for diversity). Thus, GMPEA is not a simple port of existing algorithms to the GPU; it represents an algorithmic-level innovation that unifies the design philosophy with the computational architecture, allowing it to fully leverage the parallel advantages of GPUs to provide an efficient and high-quality solution for high-timeliness CMOPs.

\begin{algorithm}[!htb]
	\caption{The framework of GMPEA} 
	\label{a1}
	\begin{algorithmic}[1]
		\Require $n$: The population size.  $t_1$: The size of neighbors $\boldsymbol{B}_{1}$. $t_2$: The size of neighbors $\boldsymbol{B}_{2}$.  $K_{max}$: The maximal number of generations. $\boldsymbol{W}$: Uniform reference vectors
		
		\Ensure $Pop_{1} \{\boldsymbol{X}_1, \boldsymbol{F}_1, \boldsymbol{C}_1\}$.
		\State $\boldsymbol{X}_{1}$, $\boldsymbol{X}_{2}$ $\gets$ Tensorized  initialization 
		\State $\boldsymbol{F}_{1}$, $\boldsymbol{F}_{2}$, $\boldsymbol{C}_{1}$, $\boldsymbol{C}_{2}$ $\gets$ Tensorized evaluate objectives
		\Statex  and constraints.
		\State $\boldsymbol{B}_{1}$, $\boldsymbol{B}_{2}$ $\gets$ Neighbours initialization
		\State $z$ $\gets$ Ideal point initialization
		\For{$k=1$ to $K_{max}$}
		\State $\boldsymbol{X}_\textit{off1}$ , $\boldsymbol{X}_\textit{off2}$ $\gets$ Tensorized Reproduction operator.
		\State $\boldsymbol{F}_\textit{off1}$ , $\boldsymbol{F}_\textit{off2}$, $\boldsymbol{C}_\textit{off1}$ , $\boldsymbol{C}_\textit{off2}$ $\gets$ Tensorized evaluate
		\Statex \quad \; objectives and constraints.
		\State $z$ $\gets$ Ideal points update
		\State $Pop_{1}\{\boldsymbol{X}_1,\boldsymbol{F}_1,\boldsymbol{C}_1\}$, $Pop_{2}\{\boldsymbol{X}_2,\boldsymbol{F}_2,\boldsymbol{C}_2\}$ $\gets$ Ten-
		\Statex \quad \; sorized environmental selection of GMPEA
		\EndFor
		\\
		\Return $Pop_{1}\{\boldsymbol{X}_1,\boldsymbol{F}_1,\boldsymbol{C}_1\}$;
	\end{algorithmic}
\end{algorithm}

\begin{algorithm}[!htb]
	\caption{Tensorized Environmental Selection of GMPEA} 
	\label{a2}
	\begin{algorithmic}[1]
		\Require Population and offspring tensors: 
		\Statex $Pop_1\{\boldsymbol{X}_1,\boldsymbol{F}_1,\boldsymbol{C}_1\}$, $Pop_2\{\boldsymbol{X}_2,\boldsymbol{F}_2,\boldsymbol{C}_2\}$,
		\Statex $Off_1\{\boldsymbol{X}_\textit{off1},\boldsymbol{F}_\textit{off1},\boldsymbol{C}_\textit{off1}\}$,$Off_2\{\boldsymbol{X}_\textit{off2},\boldsymbol{F}_\textit{off2},\boldsymbol{C}_\textit{off2}\}$.  Population size: $n$. Reference vectors: $\boldsymbol{W}$. Ideal point: $z$. Neighbour indices of two populations: $\boldsymbol{B}_{1}$, $\boldsymbol{B}_{2}$.
		
		\Ensure $Pop^\textit{next}_{1} \{\boldsymbol{X}_1, \boldsymbol{F}_1, \boldsymbol{C}_1\}$, $Pop^\textit{next}_{2} \{\boldsymbol{X}_2, \boldsymbol{F}_2, \boldsymbol{C}_2\}$.
		\State $\boldsymbol{M}_1 \gets \mathbf{0}$
		\State $\boldsymbol{M}_2 \gets \mathbf{0}$
		\State Sequence tensor: $\boldsymbol{I}_\textit{sub} \gets [i \mid i \in \mathbb{Z}, 0 \leq i < N]$.
		\Function{$OP_1$}{\{$\boldsymbol{x}_\textit{off1}$,$\boldsymbol{f}_\textit{off1}$,$\boldsymbol{c}_\textit{off1}$\},\{$\boldsymbol{x}_\textit{off2}$,$\boldsymbol{f}_\textit{off2}$,$\boldsymbol{c}_\textit{off2}$
			\}, $\boldsymbol{w}$}
		\State $\boldsymbol{g}_1$ $\gets$ $f^{\textit{pbi}}(\boldsymbol{f}_\textit{off1},\boldsymbol{w},\boldsymbol{z})$
		\State $\boldsymbol{g}_2$ $\gets$ $f^{\textit{pbi}}(\boldsymbol{f}_\textit{off2},\boldsymbol{w},\boldsymbol{z})$
		\State $\boldsymbol{cv}_1$ $\gets$ $CV(\boldsymbol{c}_\textit{off1})$
		\State $\boldsymbol{cv}_2$ $\gets$ $CV(\boldsymbol{c}_\textit{off2})$
		\State $\boldsymbol{S}_1$ $\gets$ $(\boldsymbol{g}_1 > \boldsymbol{g}_2 \land \boldsymbol{cv}_1 = \boldsymbol{cv}_2) \lor (\boldsymbol{cv}_2 <$ 
		\Statex \quad \; $\boldsymbol{cv}_1)$
		\State $\boldsymbol{S}_2$ $\gets$ $(\boldsymbol{g}_2 > \boldsymbol{g}_1)$
		
		\State \{$\boldsymbol{x}^\textit{new}_\textit{off1}$, $\boldsymbol{f}^\textit{new}_\textit{off1}$, $\boldsymbol{c}^\textit{new}_\textit{off1}$\} $\gets$ $H(\boldsymbol{S}_1)$ $\odot$ \{$\boldsymbol{x}_\textit{off2}$, $\boldsymbol{f}_\textit{off2}$, $\boldsymbol{c}_\textit{off2}$\} +  
		\Statex \quad \; $(1-H(\boldsymbol{S}_1))$ $\odot$ \{$\boldsymbol{x}_\textit{off1}$, $\boldsymbol{f}_\textit{off1}$, $\boldsymbol{c}_\textit{off1}$\}
		
		\State \{$\boldsymbol{x}^\textit{new}_\textit{off2}$, $\boldsymbol{f}^\textit{new}_\textit{off2}$, $\boldsymbol{c}^\textit{new}_\textit{off2}$\} $\gets$ $H(\boldsymbol{S}_2)$ $\odot$ \{$\boldsymbol{x}_\textit{off1}$, $\boldsymbol{f}_\textit{off1}$, $\boldsymbol{c}_\textit{off1}$\} +  
		\Statex \quad \; $(1-H(\boldsymbol{S}_1))$ $\odot$ \{$\boldsymbol{x}_\textit{off2}$, $\boldsymbol{f}_\textit{off2}$, $\boldsymbol{c}_\textit{off2}$\}
		\\
		\quad\; \Return \{$\boldsymbol{x}^\textit{new}_\textit{off1}$, $\boldsymbol{f}^\textit{new}_\textit{off1}$, $\boldsymbol{c}^\textit{new}_\textit{off1}$\}, \{$\boldsymbol{x}^\textit{new}_\textit{off2}$, $\boldsymbol{f}^\textit{new}_\textit{off2}$, $\boldsymbol{c}^\textit{new}_\textit{off2}$\}
		\EndFunction

		\Function{$OP_2$}{$\boldsymbol{b}_1$,$\boldsymbol{b}_2$,\{$\boldsymbol{x}^\textit{new}_\textit{off1}$, $\boldsymbol{f}^\textit{new}_\textit{off1}$,$\boldsymbol{c}^\textit{new}_\textit{off1}$\},\{$\boldsymbol{x}^\textit{new}_\textit{off2}$, $\boldsymbol{f}^\textit{new}_\textit{off2}$,$\boldsymbol{c}^\textit{new}_\textit{off2}$\}}
		\State $\boldsymbol{g}_\textit{old1} \gets f^{\textit{pbi}}(\boldsymbol{F}_\textit{1}[\boldsymbol{b}_1],\boldsymbol{W}[b_1],\boldsymbol{z})$
		\State $\boldsymbol{g}_\textit{old2} \gets f^{\textit{pbi}}(\boldsymbol{F}_\textit{2}[\boldsymbol{b}_2],\boldsymbol{W}[b_2],\boldsymbol{z})$
		\State $\boldsymbol{g}_\textit{new1} \gets f^{\textit{pbi}}(\boldsymbol{f}^\textit{new}_\textit{off1},\boldsymbol{W}[\boldsymbol{b}_1],\boldsymbol{z})$
		\State $\boldsymbol{g}_\textit{new2} \gets f^{\textit{pbi}}(\boldsymbol{f}^\textit{new}_\textit{off2},\boldsymbol{W}[\boldsymbol{b}_2],\boldsymbol{z})$
		\State $\boldsymbol{cv}_\textit{old1} \gets CV(\boldsymbol{C}_1[\boldsymbol{b}_1])$
		\State $\boldsymbol{cv}_\textit{new1} \gets CV(\boldsymbol{c}^\textit{new}_\textit{off1})$
		\State $\boldsymbol{S}_1$ $\gets$ $(\boldsymbol{g}_\textit{old1} > \boldsymbol{g}_\textit{new1} \land \boldsymbol{cv}_\textit{old1} = \boldsymbol{cv}_\textit{new1}) \lor (\boldsymbol{cv}_\textit{new1} <$ 
		\Statex \quad \; $\boldsymbol{cv}_\textit{old1})$
		\State $\boldsymbol{S}_2$ $\gets$ $(\boldsymbol{g}_\textit{old2} > \boldsymbol{g}_\textit{new2})$
		\State $\boldsymbol{M}_1[\boldsymbol{b}_1] \gets H(\boldsymbol{S}_1)$
		\State $\boldsymbol{M}_2[\boldsymbol{b}_2] \gets H(\boldsymbol{S}_2)$
		\State $\boldsymbol{I}_\textit{sub1} \gets \boldsymbol{M}_1 \odot (-1) + (1-\boldsymbol{M}_1) \odot \boldsymbol{I}_\textit{sub}$
		\State $\boldsymbol{I}_\textit{sub2} \gets \boldsymbol{M}_2 \odot (-1) + (1-\boldsymbol{M}_2) \odot \boldsymbol{I}_\textit{sub}$\\
		\quad\; \Return $\boldsymbol{I}_\textit{sub1}$, $\boldsymbol{I}_\textit{sub2}$
		\EndFunction
		
		\Function{$OP_3$}{$\boldsymbol{i}_1$,$\boldsymbol{i}_2$,\{$\boldsymbol{x}_1$, $\boldsymbol{f}_1$,$\boldsymbol{c}_1$\},\{$\boldsymbol{x}_2$,$\boldsymbol{f}_2$,  $\boldsymbol{c}_2$\}, $\boldsymbol{w}$}
		\State $\{\boldsymbol{x}^{'}_{1}, \boldsymbol{f}^{'}_{1}, \boldsymbol{c}^{'}_{1}\} \gets H(\boldsymbol{i}_1=-1) \odot \{\boldsymbol{X}_\textit{off1}, \boldsymbol{F}_\textit{off1}, \boldsymbol{C}_\textit{off1}\} + $
		\Statex \quad \; $(1 - H(\boldsymbol{i}_1=-1)) \odot \{\boldsymbol{x}_1, \boldsymbol{f}_1, \boldsymbol{c}_1\}$;
		
		\State $\{\boldsymbol{x}^{'}_{2}, \boldsymbol{f}^{'}_{2}, \boldsymbol{c}^{'}_{2}\} \gets H(\boldsymbol{i}_2=-1) \odot \{\boldsymbol{X}_\textit{off2}, \boldsymbol{F}_\textit{off2}, \boldsymbol{C}_\textit{off2}\} + $
		\Statex \quad \; $(1 - H(\boldsymbol{i}_2=-1)) \odot \{\boldsymbol{x}_2, \boldsymbol{f}_2, \boldsymbol{c}_2\}$;
		
		\State $p\leftarrow\operatorname{argmin}(CV(\boldsymbol{c}^{'}_{1}))$;    Multiple minimum CVs may 
		\Statex \quad \; exist. 
		\State $j_1\leftarrow\operatorname{argmin}(f^{\textit{pbi}}(\boldsymbol{f}^{'}_{1}[p],\boldsymbol{w},\boldsymbol{z}))$;
		\State $j_2\leftarrow\operatorname{argmin}(f^{\textit{pbi}}(\boldsymbol{f}^{'}_{2},\boldsymbol{w},\boldsymbol{z}))$;\\
		\quad\; \Return \{$\boldsymbol{x}_1[j_1]$,$\boldsymbol{f}_1[j_1]$,$\boldsymbol{c}_1[j_1]$\}, \{$\boldsymbol{x}_2[j_2]$,$\boldsymbol{f}_2[j_2]$,$\boldsymbol{c}_2[j_2]$\}
		\EndFunction
		\\
		\State $Off^\textit{new}_1, Off^\textit{new}_2$ $\gets$ $\mathbf{vmap}(OP_1)(Off_1, Off_1)$
		\State $\boldsymbol{I}_1$, \quad $\boldsymbol{I}_2$ $\gets$ $\mathbf{vmap}(OP_2)(\boldsymbol{B}_{1}, \boldsymbol{B}_{1}, Off^\textit{new}_1, Off^\textit{new}_2)$
		
		\State $Pop^\textit{next}_{1}, Pop^\textit{next}_{2}$ $\gets$  $\mathbf{vmap}(OP_3)(\boldsymbol{I}^{\top}_1, \boldsymbol{I}^{\top}_2, Pop_{1}, Pop_{2})$ \\
		\Return $Pop^\textit{next}_{1}, Pop^\textit{next}_{2}$
	\end{algorithmic}
\end{algorithm}

\section{Experimental Setup}
\label{s4}
This section details the experimental setup used to evaluate the performance of GMPEA, including the experimental environment, test problems, comparative algorithms, performance metrics, and statistical methods. To provide a comprehensive and in-depth analysis, we designed two experimental schemes. The first scheme assesses the algorithms' ability to solve CMOPs without time constraints, giving all algorithms a fixed, generous number of function evaluations. The second scheme addresses the high-timeliness requirements of real-world CMOPs by limiting the total solution time to 10 seconds, which evaluates the algorithms' performance under rapid solution conditions.

\subsection{Experimental Environment} 
All experiments were conducted on a single high-performance personal workstation to ensure a consistent and efficient computing environment. The specific hardware configuration is as follows:

CPU: Intel(R) Core(TM) i5-12400F. GPU: NVIDIA GeForce RTX 4080. RAM: 32GB. Programming Environment: python 3.11 and cuda 12.1.

\subsection{Test Problems}  
For our experimental evaluation, we selected several widely accepted CMOP test suites and  real-world weapon-target assignment (WTA) problems \cite{li2024knowledge}. The WTA problems are  high-timeliness problems that demand fast solutions, typically within seconds. The LIRCMOP\cite{fan2017improved} test suite, consisting of 14 problems, is specifically designed to assess an algorithm's performance when dealing with large infeasible regions. The C-DTLZ \cite{jain2013evolutionary} and DC-DTLZ \cite{li2018two} test suites are well-known constrained optimization problems that provide various types of complex constraints for multiobjective optimizers.

\subsection{Algorithms and Parameter Settings}  
To validate the effectiveness of the proposed GMPEA, we conducted comprehensive comparisons with several state-of-the-art and representative algorithms. Given that all algorithms require implementation within a GPU parallel environment, we carefully selected methods that are both architecturally straightforward and extensively cited in the literature, taking into account the complexities associated with code migration, system compatibility, and suitability for tensorized restructuring. From each major category of CMOEAs, we selected one representative method: c-NSGA-II (feasibility-prioritized approach), PPS (constraint relaxation-based), CMOEA-MS (multi-stage approach), CCMO (multi-population approach), and EMCMO (multi-task approach). All algorithms utilize the robust tensorized implementations for initialization, mating selection, evolutionary operators, and function evaluation that are provided by the EvoX platform. Specifically, c-NSGA-II is adapted from TensoredNSGA-II \cite{huang2024evox}, while the main framework of PPS is based on TensoredMOEA/D \cite{huang2024evox} and incorporates the environmental selection of TensoredNSGA-II for its archive set updates. Each algorithm is run independently 20 times across all test problems. In GMPEA, we use both large and small neighborhoods. The large neighborhood is set to a size of $t_2: 2*T$, while the small one is set to $t_1: T/2$. The value of $T$ follows the common practice of most decomposition-based MOEAs and is set to 10 \cite{zhang2007moea}.

The main experimental parameters are set as follows:
﻿
\subsubsection{Population Size and Computational Resources}  
In both the fixed-evaluation and fixed-time experiments, the population size for all algorithms was uniformly set to 1,000.  For the fixed-evaluation experiments, the maximum number of evaluations for all algorithms across all problems was uniformly set to $10^6$. In the fixed-time experiments, the solution time was strictly limited to 10 seconds. When an algorithm's runtime reached 10 seconds, it was immediately terminated, and its current solution set was output.  
﻿
\subsubsection{Decision Variables and Objectives}    
For C1-DTLZ1, DC1-DTLZ1, DC2-DTLZ1, and DC3-DTLZ1, the number of decision variables ($d$) was set to 7. For the remaining problems in the C-DTLZ and DC-DTLZ test suites, $d$ was set to 12. For LIR-CMOPs, $d$ was set to 30. The number of objectives ($m$) was set to 3 for LIRCMOP13, LIR-CMOP14, C-DTLZ, and DC-DTLZ, while $m$ was set to 2 for the remaining CMOPs.

\subsubsection{Evolutionary Operators} 
All algorithms use the same evolutionary operators as their respective counterparts in related literature. Crucially, all these operators are implemented in their tensorized versions to leverage GPU parallelism. For the C-DTLZ and DC-DTLZ test suites, all algorithms employ the simulated binary crossover (SBX) and polynomial mutation (PM). The crossover probability for SBX [40] is set to 1, and the mutation probability for PM [41] is set to 1/D, with a distribution index of 20 for both operators. For the LIRCMOPs, all algorithms use Differential Evolution (DE), with the CR and F parameters set to 1 and 0.5, respectively.

\subsection{Performance Metrics} 

To comprehensively evaluate the algorithms' convergence and diversity, we adopted two key performance metrics: Inverted Generational Distance (IGD) \cite{bosman2003balance} and Hypervolume (HV) \cite{while2006faster}.

For the C-DTLZ, DC-DTLZ, and LIR-CMOP test suites, we utilized the IGD metric. For the WTA problems, the HV metric is used. The IGD is a comprehensive metric that measures the average distance between the solution set obtained by an algorithm and a true or approximate Pareto front, thereby reflecting both convergence and distribution. A lower IGD value indicates better algorithm performance. The formula is as follows:
﻿
\begin{equation}  
	IGD(P, P') = \frac{\sum_{\mathbf{v} \in P'} d(\mathbf{v}, P)}{|P'|},  
\end{equation}  
﻿
where \(d(\mathbf{v}, P)\) represents the distance from a solution \(\mathbf{v}\) to the set of solutions \(P\).  
﻿
The HV [42] measures both the convergence and diversity of a solution set by calculating the volume of the objective space enclosed by the solutions and a predefined reference point. Since HV does not require prior knowledge of the true Pareto front, it is an invaluable metric for evaluating algorithm performance on real-world problems where the true PF is often unknown. A higher HV value indicates better algorithm performance. Its definition is as follows: 
﻿
\begin{equation}  
	HV(P) = V\left(\bigcup_{x \in P}\left[f_{1}(x), z_{1}\right] \times \ldots \times \left[f_{m}(x), z_{m}\right]\right),  
\end{equation}  
﻿
where \(V(\cdot)\) is the Lebesgue measure, \(m\) denotes the number of objectives, and \(\mathbf{z} = (z_{1}, \ldots, z_{m})^{T}\) is a reference point in the objective space that can dominate the  Pareto Front.  
﻿

Additionally, the Wilcoxon rank-sum test was employed in this study to analyze the results at a significance level of 0.05. The symbols “+”, “-”, and ``$= $'' indicate whether the algorithm's performance was significantly better, significantly worse, or statistically similar to its LHO versions, respectively. 

\section{Experimental Results and Discussion}
\label{s5}
This section provides a comprehensive analysis of the performance of GMPEA against several other state-of-the-art CMOEAs. Our discussion is based on two experimental schemes (fixed number of function evaluations and fixed solving time), as well as results from ablation studies and varying population sizes. All findings are rigorously validated through statistical methods to ensure the robustness of our conclusions.
\subsection{Results under Fixed Evaluations}
Tables \ref{t1} and \ref{t2} present the running times (in seconds) and IGD results, respectively, for GMPEA and five comparison algorithms on the C-DTLZ, DC-DTLZ, and LIRCMOP test suites under a fixed number of function evaluations. Several key observations can be made from the data.

\begin{figure*}[!htbp]
	\centering
	\footnotesize
	\begin{minipage}{8.9cm}
		\centering
		\includegraphics[width=8.6cm]{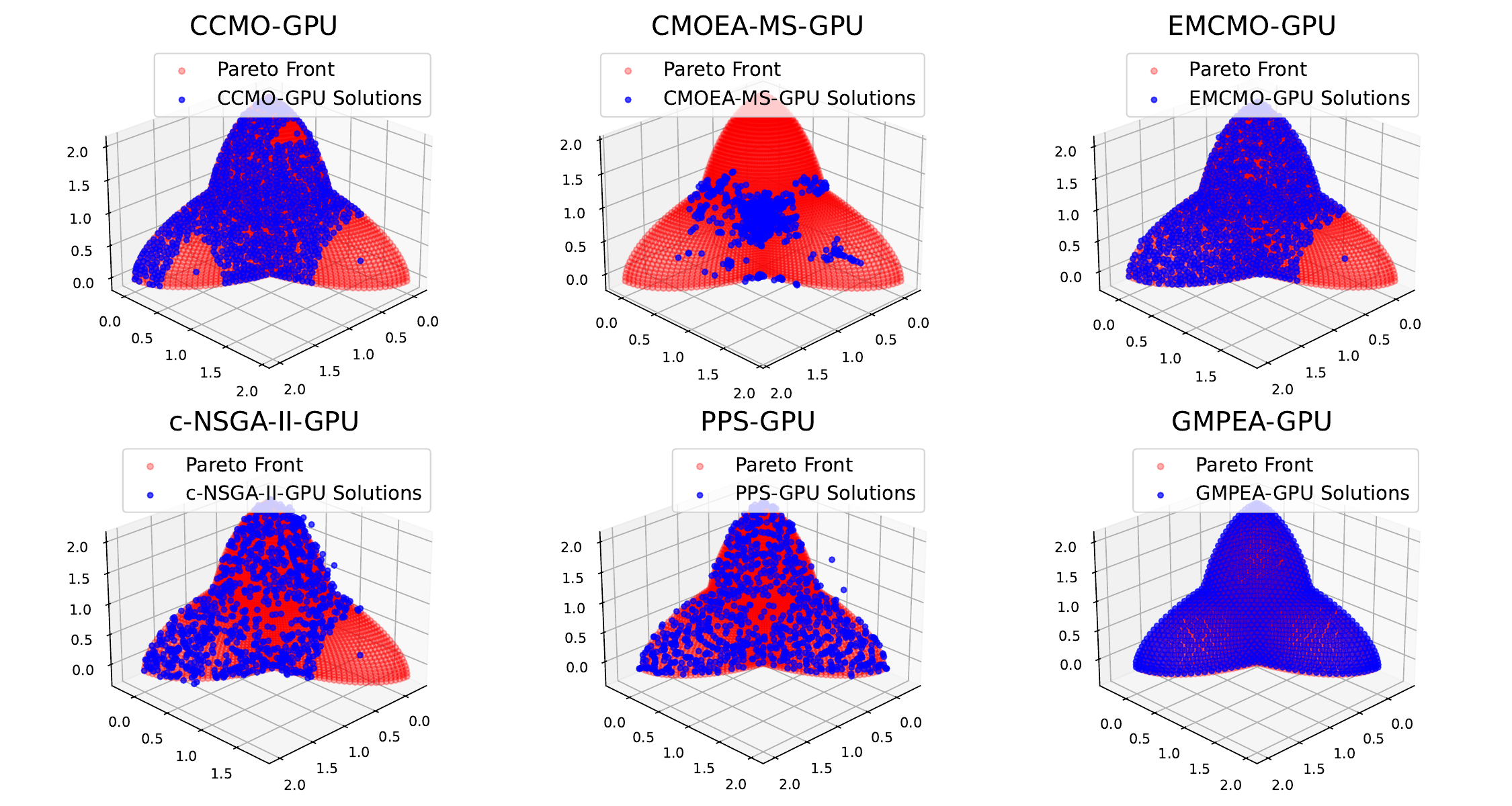}
		\text{(a) C3-DTLZ4}
	\end{minipage}
	\vspace{5pt}
	\begin{minipage}{8.9cm}
		\centering
		\includegraphics[width=8.6cm]{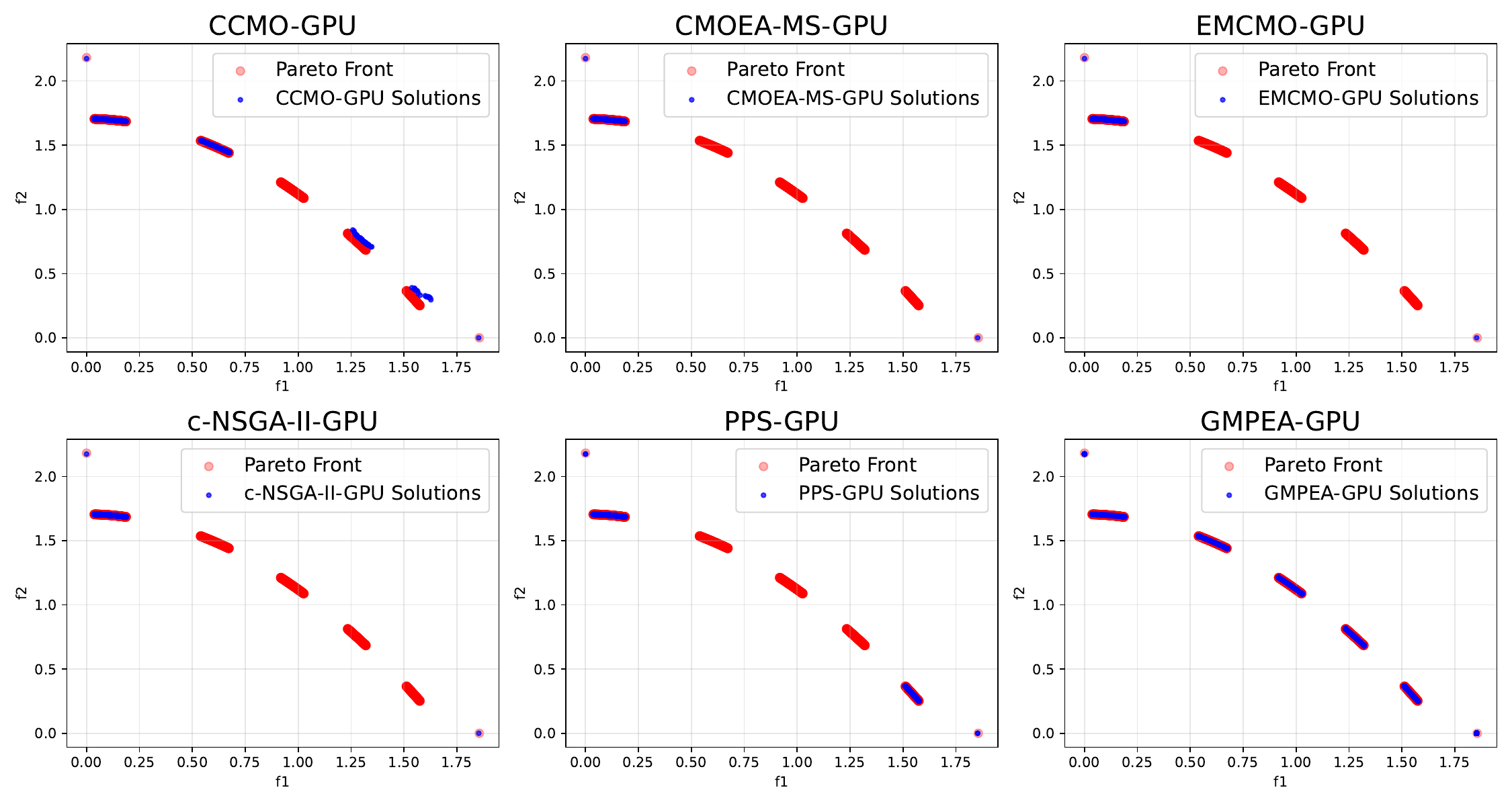}
		\text{(b) LIRCMOP9}
	\end{minipage}
	\centering
	\caption{Populations obtained by all algorithms on an NVIDIA RTX 4080 after a fixed number of function evaluations for problems (a) C3-DTLZ4 and (b) LIRCMOP9.}
	\label{f3}
\end{figure*}

First, in terms of solution quality, GMPEA is competitive with all other algorithms, achieving the best or a comparable IGD value on most problems. Notably, for complex problems like C-DTLZ3 and LIRCMOP9, GMPEA's IGD values are significantly superior to all competing algorithms. Fig. \ref{f3}, which shows the final populations obtained on C3-DTLZ4 and LIRCMOP9, reveals that only GMPEA achieved a population with both good diversity and convergence. This demonstrates that even with ample computational resources, GMPEA's decomposition-based dual-population co-evolutionary mechanism effectively balances convergence and diversity to handle complex constraints. This aligns with our design philosophy where $Pop_1$ (enhanced local exploration with constraint consideration) and $Pop_2$ (enhanced global convergence without constraints) work collaboratively, proving that this strategy achieves efficient search without compromising solution quality.

Second, in terms of computational efficiency, GMPEA exhibits a decisive advantage. As shown in Table \ref{t1}, GMPEA's average runtime is significantly lower than that of all other algorithms across all test problems. For example, on the C1-DTLZ1 problem, GMPEA's average runtime is only 10.24 seconds, while the next closest algorithm, PPS-GPU, takes 17.73 seconds. Other algorithms, such as CCMO-GPU and EMCMO-GPU, require as much as 548.52 and 650.91 seconds, respectively. This substantial time difference (a speedup of tens to hundreds of times) provides strong evidence that GMPEA's fully tensorized and parallel design offers immense performance gains in a GPU environment. It also validates our central argument that most existing CMOEAs are hindered by their complex environmental selection strategies (e.g., serially dependent truncation or non-dominated sorting), which create a severe computational bottleneck and prevent them from fully leveraging GPU parallelism.

\begin{figure*}[!htbp]
	\centering
	\footnotesize
	\begin{minipage}{5.8cm}
		\centering
		\includegraphics[width=5.9cm]{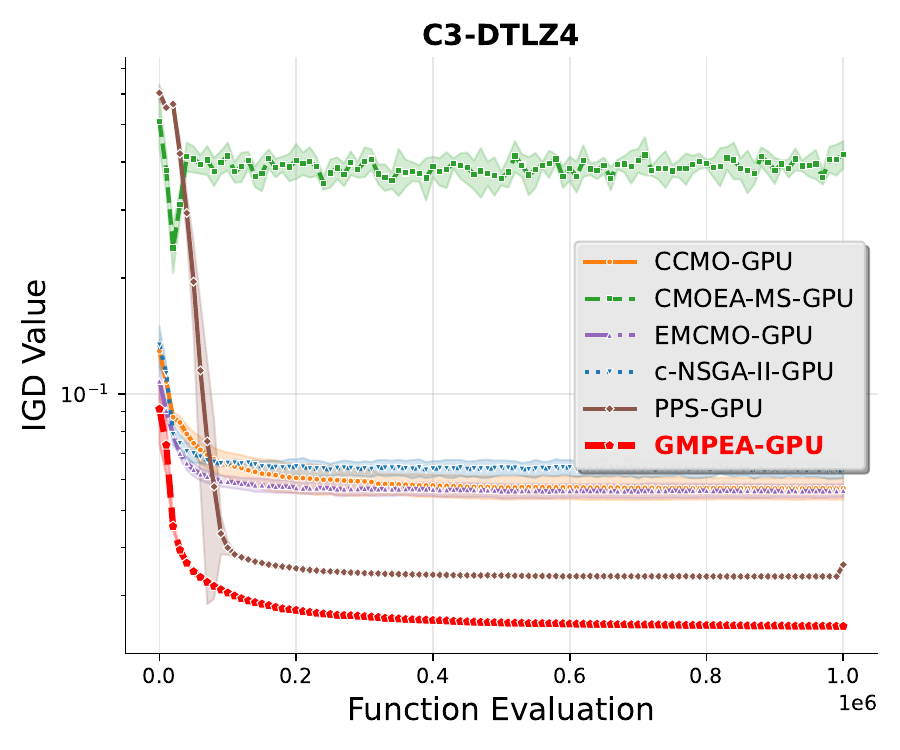}
		\text{(a)}
	\end{minipage}
	\vspace{5pt}
	\begin{minipage}{5.8cm}
		\centering
		\includegraphics[width=5.9cm]{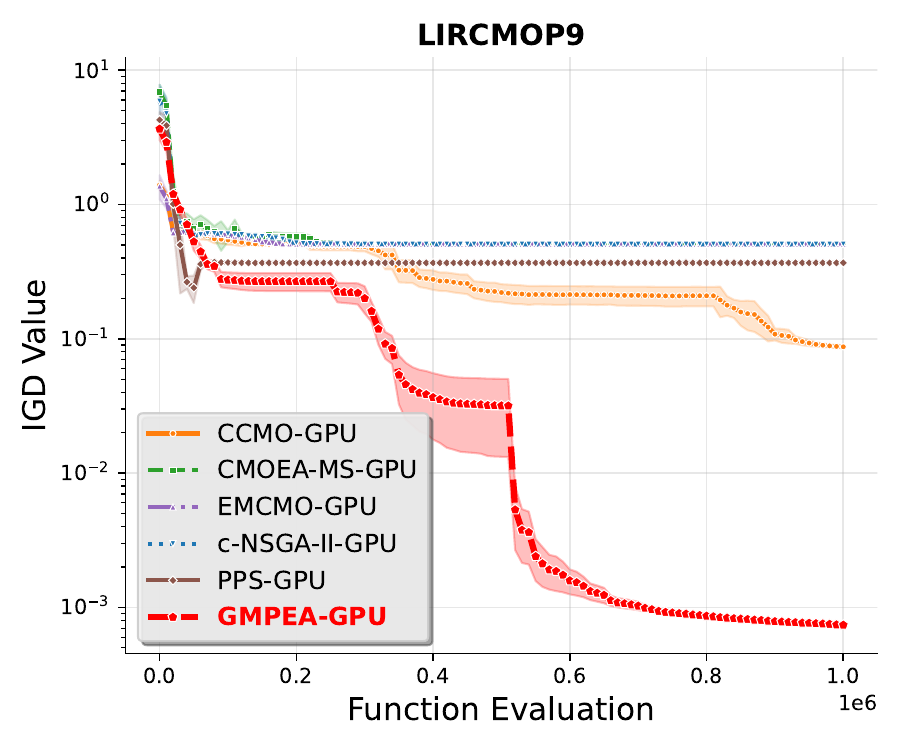}
		\text{(b)}
	\end{minipage}
	\begin{minipage}{5.8cm}
		\centering
		\includegraphics[width=5.9cm]{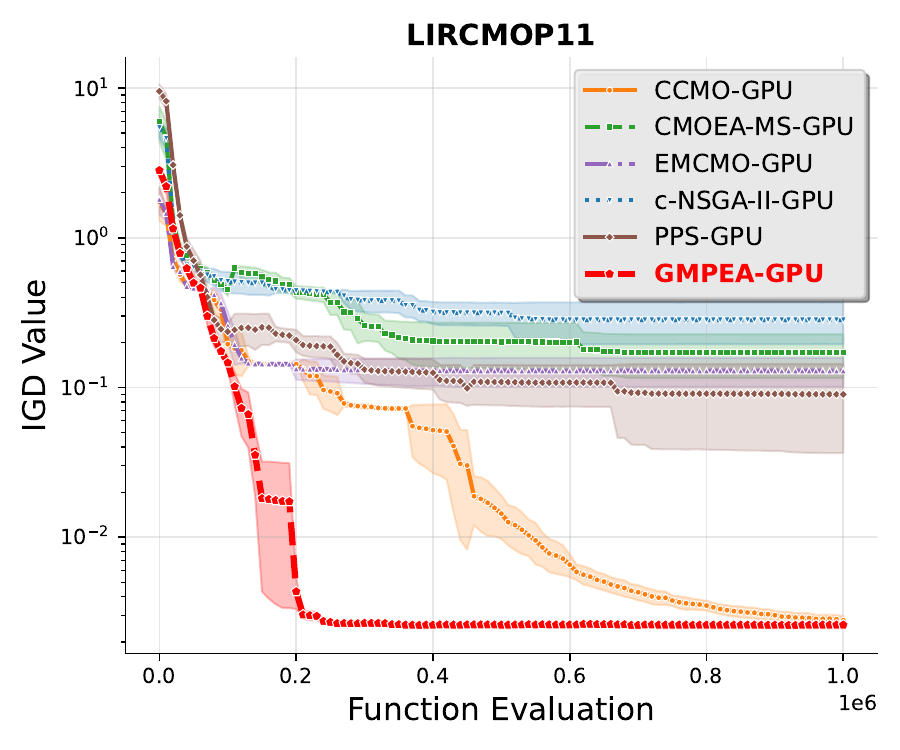}
		\text{(c)}
	\end{minipage}
	\centering
	
	\caption{Average IGD convergence curves for all algorithms on problems (a) C3-DTLZ4, (b) LIRCMOP9, and (c) LIRCMOP11 under a fixed number of function evaluations. Shaded areas represent the upper and lower standard deviation ranges.}
	\label{f4}
\end{figure*}

Fig. \ref{f4} further shows the IGD convergence curves of GMPEA and other algorithms on problems like C3-DTLZ4, LIRCMOP9, and LIRCMOP11. The curves clearly indicate that GMPEA's convergence speed far surpasses all competitors. In the early stages of evolution, GMPEA is able to quickly find and continuously optimize a high-quality set of solutions. This further confirms GMPEA's superior efficiency.
\begin{table*}[htbp]
	\centering
	\setlength{\tabcolsep}{2pt}
	\caption{Comparison of Average Running Time (seconds) and Standard Deviation for Different Algorithms on C-DTLZ, DC-DTLZ, and LIRCMOP Test Suites under a Fixed Number of Function Evaluations.}
	{\fontsize{7.58pt}{10.4pt}\selectfont
	\begin{tabular}{p{5.28em}|c|c|c|c|c|c}
		\toprule
		\multicolumn{1}{c|}{\textbf{Problem}} & \textbf{c-NSGA-II-GPU} & \textbf{PPS-GPU} & \textbf{CMOEA-MS-GPU} & \textbf{CCMO-GPU} & \textbf{EMCMO-GPU} & \textbf{GMPEA-GPU} \\
		\midrule
		\multicolumn{1}{c|}{C1-DTLZ1} & 238.47888 ± 1.48194 - & 17.73379 ± 0.72517 - & 315.24291 ± 7.97429 - & 548.52201 ± 42.28362 - & 650.91588 ± 8.91329 - & \textbf{10.23567 ± 0.28569} \\
		\multicolumn{1}{c|}{C1-DTLZ3} & 51.61664 ± 1.19665 - & 18.39492 ± 0.07148 - & 178.98168 ± 14.17076 - & 434.04770 ± 20.34378 - & 495.21233 ± 3.16908 - & \textbf{10.26443 ± 0.07488} \\
		\multicolumn{1}{c|}{C2-DTLZ2} & 71.60097 ± 1.21704 - & 17.33626 ± 0.22202 - & 353.85023 ± 41.16616 - & 477.03123 ± 103.42011 - & 776.03387 ± 17.93070 - & \textbf{10.65504 ± 0.18951} \\
		\multicolumn{1}{c|}{C3-DTLZ4} & 194.18388 ± 1.50033 - & 18.89075 ± 0.88966 - & 1430.04837 ± 296.35658 - & 188.84873 ± 10.95819 - & 273.59343 ± 1.47224 - & \textbf{10.18076 ± 0.10054} \\
		\multicolumn{1}{c|}{DC1-DTLZ1} & 33.83101 ± 0.39929 - & 15.65991 ± 0.68687 - & 677.96603 ± 103.33705 - & 213.27600 ± 12.32541 - & 430.28852 ± 5.22384 - & \textbf{10.12380 ± 0.11589} \\
		\multicolumn{1}{c|}{DC1-DTLZ3} & 31.82392 ± 0.85056 - & 18.31707 ± 0.47614 - & 247.62499 ± 2.14114 - & 283.64233 ± 59.30380 - & 427.19351 ± 2.51654 - & \textbf{10.28650 ± 0.46213} \\
		\multicolumn{1}{c|}{DC2-DTLZ1} & 255.74353 ± 12.50723 - & 16.83502 ± 0.50640 - & 337.62999 ± 7.31797 - & 384.97826 ± 55.56420 - & 590.00848 ± 7.27185 - & \textbf{10.47364 ± 0.21256} \\
		\multicolumn{1}{c|}{DC2-DTLZ3} & 281.59725 ± 27.22730 - & 15.82118 ± 0.70372 - & 460.13949 ± 4.96710 - & 282.67993 ± 58.56134 - & 488.20339 ± 2.78281 - & \textbf{10.25141 ± 0.17372} \\
		\multicolumn{1}{c|}{DC3-DTLZ1} & 204.20611 ± 22.00395 - & 19.33008 ± 0.99987 - & 549.57462 ± 382.59382 - & 227.28114 ± 3.60461 - & 471.30861 ± 5.16455 - & \textbf{10.13329 ± 0.10112} \\
		\multicolumn{1}{c|}{DC3-DTLZ3} & 229.09556 ± 28.43363 - & 17.13582 ± 0.74479 - & 494.75099 ± 19.66382 - & 180.49364 ± 3.09170 - & 384.77846 ± 6.13230 - & \textbf{10.58180 ± 0.14012} \\
		\midrule
		\multicolumn{1}{c|}{LIRCMOP1} & 187.69544 ± 1.37127 - & 31.04101 ± 13.41883 - & 114.41290 ± 0.40404 - & 42.74383 ± 7.35598 - & 40.10856 ± 3.27571 - & \textbf{9.15426 ± 0.51159} \\
		\multicolumn{1}{c|}{LIRCMOP2} & 172.18139 ± 1.79268 - & 39.77512 ± 14.10570 - & 114.04921 ± 0.27424 - & 33.84156 ± 0.82300 - & 24.90663 ± 1.39816 - & \textbf{8.85572 ± 0.04529} \\
		\multicolumn{1}{c|}{LIRCMOP3} & 278.49950 ± 1.79250 - & 47.07949 ± 3.81363 - & 113.68181 ± 0.11495 - & 45.14922 ± 0.59878 - & 29.83557 ± 0.36911 - & \textbf{8.52713 ± 0.11384} \\
		\multicolumn{1}{c|}{LIRCMOP4} & 306.60560 ± 27.69743 - & 35.82338 ± 2.94997 - & 113.40018 ± 0.05842 - & 52.74921 ± 2.12532 - & 35.57736 ± 0.55201 - & \textbf{9.31727 ± 0.47265} \\
		\multicolumn{1}{c|}{LIRCMOP5} & 272.91378 ± 101.38844 - & 13.30443 ± 0.08159 - & 217.62649 ± 26.13317 - & 153.97157 ± 42.19537 - & 192.37764 ± 43.04488 - & \textbf{9.75697 ± 0.26524} \\
		\multicolumn{1}{c|}{LIRCMOP6} & 232.10693 ± 20.69437 - & 15.12131 ± 0.06768 - & 192.56122 ± 4.47662 - & 260.27641 ± 14.79939 - & 354.35177 ± 19.25766 - & \textbf{10.06565 ± 0.56367} \\
		\multicolumn{1}{c|}{LIRCMOP7} & 274.65742 ± 23.49772 - & 16.21726 ± 0.23701 - & 116.15609 ± 0.05601 - & 58.86825 ± 9.44630 - & 65.19456 ± 17.28645 - & \textbf{10.03455 ± 0.65017} \\
		\multicolumn{1}{c|}{LIRCMOP8} & 199.38441 ± 8.70734 - & 15.06032 ± 2.43221 - & 116.14169 ± 0.12797 - & 105.01897 ± 4.64340 - & 160.27908 ± 25.57469 - & \textbf{9.83261 ± 0.24941} \\
		\multicolumn{1}{c|}{LIRCMOP9} & 121.46184 ± 5.97339 - & 18.08292 ± 1.19162 - & 144.80980 ± 1.79580 - & 127.99538 ± 5.41945 - & 169.45972 ± 8.14509 - & \textbf{9.75009 ± 0.20959} \\
		\multicolumn{1}{c|}{LIRCMOP10} & 180.78008 ± 7.81886 - & 16.04828 ± 1.70294 - & 416.03192 ± 65.42170 - & 256.40907 ± 8.08096 - & 351.20078 ± 13.05577 - & \textbf{10.00912 ± 0.20837} \\
		\multicolumn{1}{c|}{LIRCMOP11} & 190.02992 ± 33.76381 - & 21.70217 ± 0.52674 - & 210.13109 ± 2.98250 - & 418.94399 ± 17.17348 - & 421.12644 ± 26.16207 - & \textbf{10.35566 ± 0.53343} \\
		\multicolumn{1}{c|}{LIRCMOP12} & 268.73617 ± 34.43367 - & 19.37454 ± 0.50893 - & 134.73322 ± 41.49307 - & 80.39956 ± 10.04930 - & 216.55831 ± 134.74652 - & \textbf{9.37201 ± 0.34171} \\
		\multicolumn{1}{c|}{LIRCMOP13} & 16.70203 ± 1.00778 - & 18.46330 ± 0.20054 - & 421.71073 ± 0.72847 - & 1983.26528 ± 168.69257 - & 1554.75641 ± 134.35347 - & \textbf{11.05786 ± 0.51696} \\
		\multicolumn{1}{c|}{LIRCMOP14} & 16.92585 ± 0.67011 - & 15.57124 ± 0.48841 - & 419.31379 ± 2.68073 - & 1816.45656 ± 17.40519 - & 1366.75491 ± 6.05003 - & \textbf{10.16641 ± 0.13689} \\
		\midrule
		Wilcoxon-Test(+/-/=) & 0/14/0 & 0/14/0 & 0/14/0 & 0/14/0 & 0/14/0 &  \\
		\midrule
		\textbf{Average Time } & 191.45705 ± 88.72509 & 20.75498 ± 8.61922 & 328.77372 ± 285.46420 & 360.70374 ± 496.89974 & 415.41767 ± 383.74894 & \textbf{9.97673 ± 0.58228} \\
		\bottomrule
	\end{tabular}%
	}
	\label{t2}%
\end{table*}%
  
\begin{table*}[htbp]
	\centering
    \setlength{\tabcolsep}{3pt}
	\caption{Comparison of Average IGD and Standard Deviation for Different Algorithms on C-DTLZ, DC-DTLZ, and LIRCMOP Test Suites under a Fixed Number of Function Evaluations.}
	\begin{tabular}{p{5.28em}|c|c|c|c|c|c}
		\toprule
		\multicolumn{1}{c|}{\textbf{Problem}} & \textbf{c-NSGA-II-GPU} & \textbf{PPS-GPU} & \textbf{CMOEA-MS-GPU} & \textbf{CCMO-GPU} & \textbf{EMCMO-GPU} & \textbf{GMPEA-GPU} \\
		\midrule
		\multicolumn{1}{c|}{C1-DTLZ1} & 0.00850  ±  0.00008 -  & 0.00899 ± 0.00013 - & 0.00613 ± 0.00002 - & 0.00610 ± 0.00001 - & 0.00601 ± 0.00004 - & \textbf{0.00571 ± 0.00003} \\
		\multicolumn{1}{c|}{C1-DTLZ3} & 0.02169 ± 0.00024 - & 0.02262 ± 0.00033 - & 0.01668 ± 0.00021 - & 0.01612 ± 0.00002 - & 0.01640 ± 0.00006 - & \textbf{0.01515 ± 0.00001 } \\
		\multicolumn{1}{c|}{C2-DTLZ2} & 0.01636 ± 0.00027 -  & 0.01740 ± 0.00014 - & 0.01234 ± 0.00010 + & 0.01221 ± 0.00007 +  & \textbf{0.01203 ± 0.00021 + } & 0.01364 ± 0.00001  \\
		\multicolumn{1}{c|}{C3-DTLZ4} & 0.06345 ± 0.00295 -  & 0.03611 ± 0.00021 - & 0.41884 ± 0.03446 - & 0.05695 ± 0.00369 -  & 0.05622 ± 0.00222 - & \textbf{0.02501 ± 0.00009 } \\
		\multicolumn{1}{c|}{DC1-DTLZ1} & 0.00407 ± 0.00004 -  & 0.00640 ± 0.00029 - & 0.00401 ± 0.00124 + & 0.00316 ± 0.00004 +  & \textbf{0.00313 ± 0.00003 +} & 0.00533 ± 0.00001  \\
		\multicolumn{1}{c|}{DC1-DTLZ3} & 0.01394 ± 0.00032 - & 0.01600 ± 0.00025 - & 0.00965 ± 0.00015 +  & 0.00976 ± 0.00010 + & \textbf{0.00975 ± 0.00016 +} & 0.01404 ± 0.00001  \\
		\multicolumn{1}{c|}{DC2-DTLZ1} & 139.87903 ± 29.48462 -  & 0.00897 ± 0.00010 - & 0.00614 ± 0.00001 - & 0.00611 ± 0.00002 - & 0.00605 ± 0.00001 - & \textbf{0.00571 ± 0.00003} \\
		\multicolumn{1}{c|}{DC2-DTLZ3} & 84.51849 ± 23.55023 -  & 0.02225 ± 0.00029 - & 0.01630 ± 0.00004 - & 0.01614 ± 0.00004 -  & 0.01640 ± 0.00006 - & \textbf{0.01515 ± 0.00002} \\
		\multicolumn{1}{c|}{DC3-DTLZ1} & 0.00253 ± 0.00005 - & 0.00411 ± 0.00011 = & 0.00314 ± 0.00256 + & \textbf{0.00177 ± 0.00004 + } & 0.00179 ± 0.00006 + & 0.00465 ± 0.00001  \\
		\multicolumn{1}{c|}{DC3-DTLZ3} & 0.11779 ± 0.21928 - & 0.01341 ± 0.00031 = & 0.02187 ± 0.03276 - & 0.00556 ± 0.00014 + & \textbf{0.00552 ± 0.00010 +} & 0.01350 ± 0.00001  \\
		\midrule
		\multicolumn{1}{c|}{LIRCMOP1} & 0.00702 ± 0.00026 - & 0.03092 ± 0.05583 = & 0.00713 ± 0.00063 - & 0.00687 ± 0.00045 - & 0.00691 ± 0.00020 - & \textbf{0.00294 ± 0.00071 } \\
		\multicolumn{1}{c|}{LIRCMOP2} & 0.00648 ± 0.00014 - & \textbf{0.00183 ± 0.00027 =} & 0.00668 ± 0.00055 - & 0.00528 ± 0.00007 - & 0.00557 ± 0.00022 - & 0.00209 ± 0.00008  \\
		\multicolumn{1}{c|}{LIRCMOP3} & 0.00509 ± 0.00030 - & 0.01020 ± 0.01147 - & 0.00573 ± 0.00016 - & 0.00475 ± 0.00076 = & 0.00459 ± 0.00030 - & \textbf{0.00315 ± 0.00116 } \\
		\multicolumn{1}{c|}{LIRCMOP4} & 0.00581 ± 0.00051 - & 0.02970 ± 0.03086 - & 0.00624 ± 0.00017 - & 0.00533 ± 0.00032 - & 0.00523 ± 0.00020 - & \textbf{0.00470 ± 0.00188 } \\
		\multicolumn{1}{c|}{LIRCMOP5} & 0.30866 ± 0.19943 - & 0.00313 ± 0.00053 - & 0.16638 ± 0.02425 - & 0.00733 ± 0.00511 - & 0.08419 ± 0.01874 - & \textbf{0.00113 ± 0.00004 } \\
		\multicolumn{1}{c|}{LIRCMOP6} & 0.61103 ± 0.36694 - & 0.00310 ± 0.00070 - & 0.38883 ± 0.03411 - & 0.00464 ± 0.00088 - & 0.13942 ± 0.15922 - & \textbf{0.00111 ± 0.00003 } \\
		\multicolumn{1}{c|}{LIRCMOP7} & 0.04275 ± 0.03400 - & 0.08473 ± 0.01115 - & 0.03423 ± 0.02108 - & \textbf{0.00126 ± 0.00006 +} & 0.00470 ± 0.00366 + & 0.03643 ± 0.03701  \\
		\multicolumn{1}{c|}{LIRCMOP8} & 0.19579 ± 0.23531 - & 0.04188 ± 0.03599 - & 0.09451 ± 0.18268 - & \textbf{0.00124 ± 0.00003 +} & 0.00227 ± 0.00041 + & 0.03290 ± 0.03996  \\
		\multicolumn{1}{c|}{LIRCMOP9} & 0.50586 ± 0.00005 - & 0.36740 ± 0.00004 - & 0.48444 ± 0.04278 - & 0.13275 ± 0.04585 - & 0.50587 ± 0.00006 - & \textbf{0.00075 ± 0.00005 } \\
		\multicolumn{1}{c|}{LIRCMOP10} & 0.22359 ± 0.01740 - & 0.09208 ± 0.08104 - & 0.00309 ± 0.00018 - & 0.00430 ± 0.00028 - & 0.21372 ± 0.00023 - & \textbf{0.00072 ± 0.00001 } \\
		\multicolumn{1}{c|}{LIRCMOP11} & 0.22823 ± 0.06932 - & 0.04809 ± 0.05730 - & 0.22676 ± 0.06801 - & 0.00266 ± 0.00010 = & 0.12928 ± 0.02842 - & \textbf{0.00254 ± 0.00004 } \\
		\multicolumn{1}{c|}{LIRCMOP12} & 0.12823 ± 0.00393 - & 0.00310 ± 0.00031 - & 0.05830 ± 0.01107 - & 0.00370 ± 0.00054 - & 0.05145 ± 0.02474 - & \textbf{0.00275 ± 0.00004 } \\
		\multicolumn{1}{c|}{LIRCMOP13} & 0.93709 ± 0.46857 - & 0.04628 ± 0.00184 - & 1.72348 ± 0.07136 - & 2.19255 ± 0.24844 - & 2.49075 ± 0.40404 - & \textbf{0.03361 ± 0.00020 } \\
		\multicolumn{1}{c|}{LIRCMOP14} & 0.31077 ± 0.15434 - & 0.04387 ± 0.00046 - & 1.63061 ± 0.05655 - & 2.10351 ± 0.29675 - & 2.49131 ± 0.46881 - & \textbf{0.02700 ± 0.00004 } \\
		\midrule
		Wilcoxon-Test(+/-/=) & 0/24/0 & 0/20/4 & 4/20/0 & 7/15/2 & 7/17/0 &  \\
		\bottomrule
	\end{tabular}%
	\label{t3}%
\end{table*}%

\subsection{Results under Fixed Time}
To evaluate the algorithms' ability to handle time-sensitive CMOPs, we conducted experiments with a fixed solving time of 10 seconds. Table \ref{t3} summarizes the IGD performance of all algorithms under this strict time limit.

The data clearly shows that under this time constraint, GMPEA's performance is far superior to that of all competing algorithms. On every test problem, GMPEA achieved the lowest IGD value, outperforming the others by a wide margin. For instance, on the LIRCMOP9 problem, GMPEA's IGD value was a mere 0.00075, whereas the second-best algorithm, CCMO-GPU, only managed an IGD value of 0.61967. This indicates that within a very short timeframe, other algorithms are unable to complete a sufficient number of evolutionary iterations due to their low computational efficiency. This severely restricts their search capabilities and prevents them from finding high-quality solutions. In contrast, GMPEA's highly efficient parallel design allows it to perform more iterations within the limited time, leading to a much better solution set.

\begin{figure*}[!htbp]
	\centering
	\footnotesize
	\begin{minipage}{8.9cm}
		\centering
		\includegraphics[width=8.6cm]{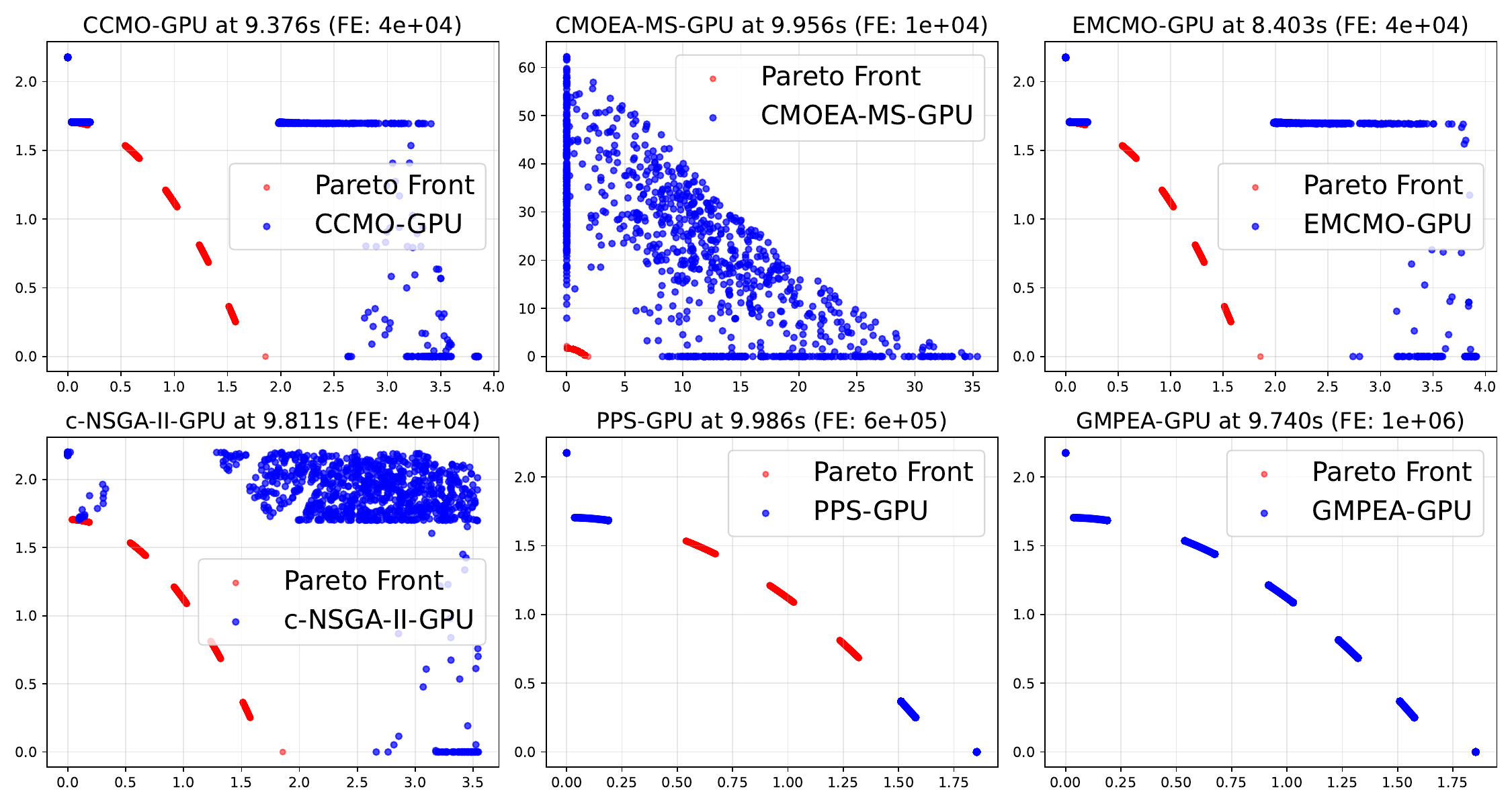}
		\text{(a) LIRCMOP9}
	\end{minipage}
	\vspace{5pt}
	\begin{minipage}{8.9cm}
		\centering
		\includegraphics[width=8.6cm]{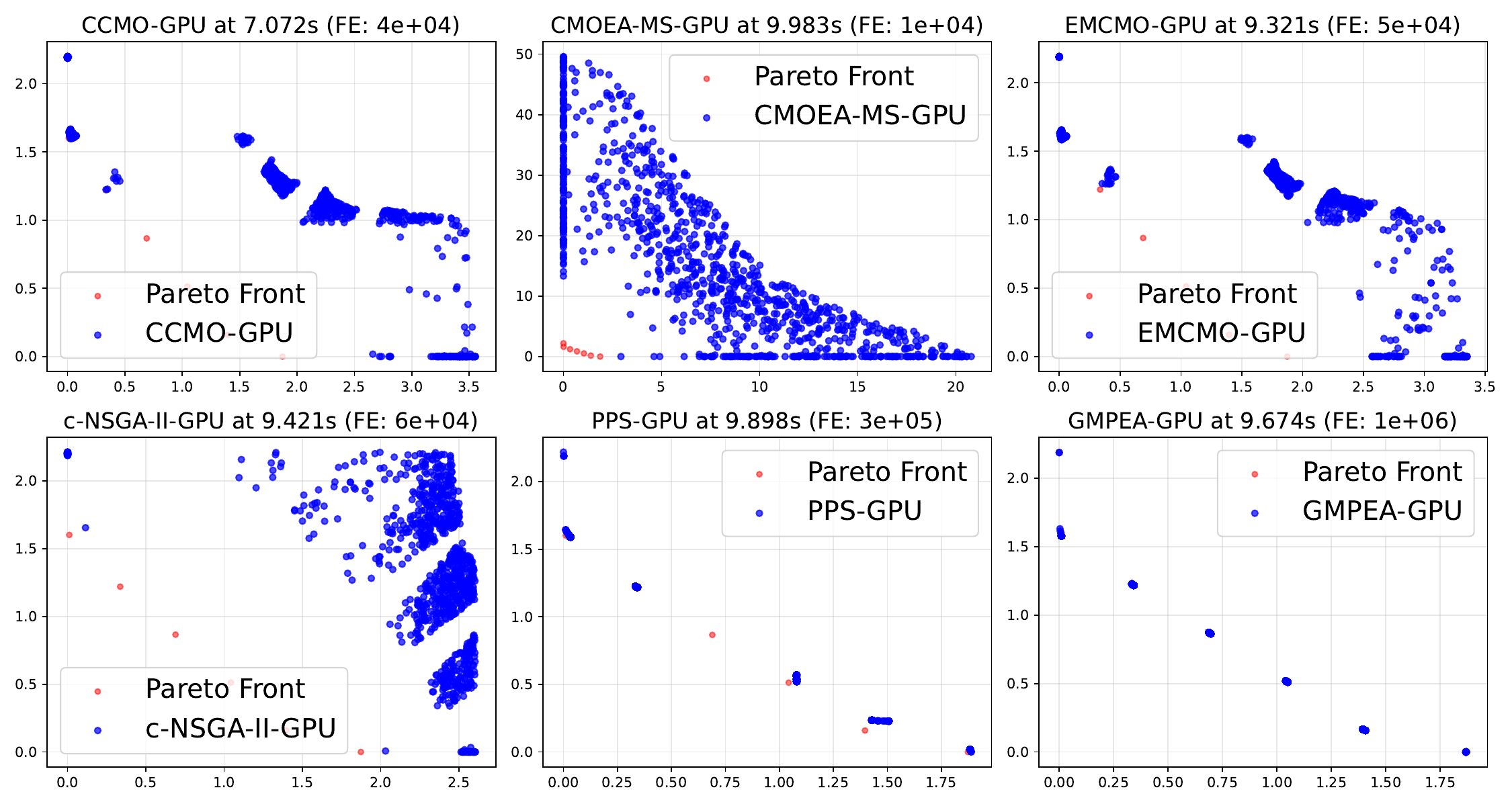}
		\text{(b) LIRCMOP11}
	\end{minipage}
	\centering
	\caption{Final populations obtained by all algorithms on an NVIDIA RTX 4080 after a fixed solving time of 10 seconds for problems (a) C3-DTLZ4 and (b) LIRCMOP9.}
	\label{f5}
\end{figure*}

\begin{figure*}[!htbp]
	\centering
	\footnotesize
	\begin{minipage}{5.8cm}
		\centering
		\includegraphics[width=5.9cm]{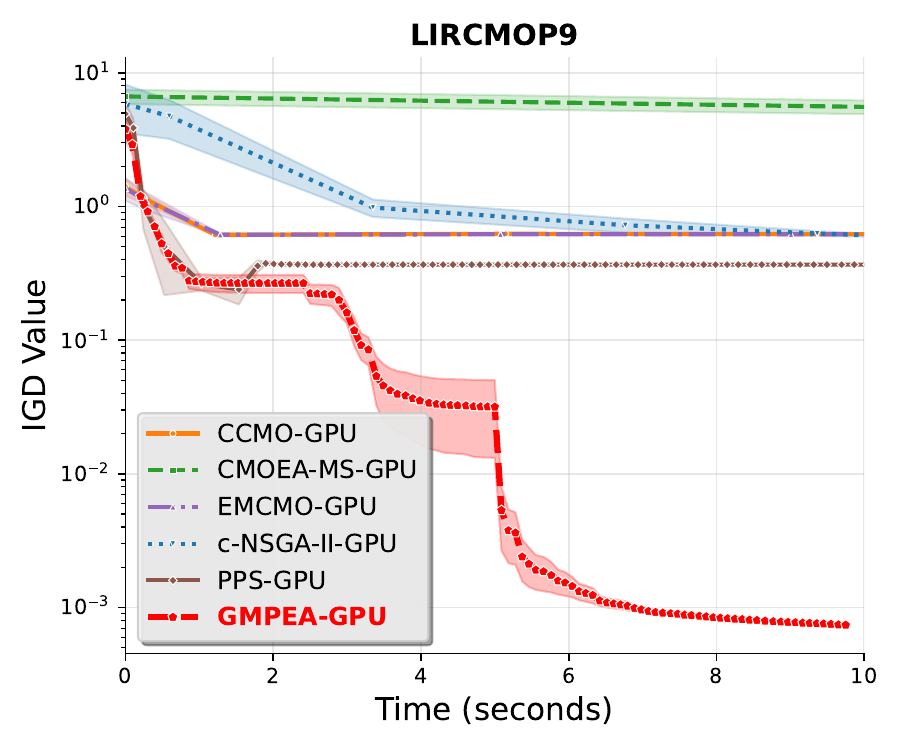}
		\text{(a)}
	\end{minipage}
	\vspace{5pt}
	\begin{minipage}{5.8cm}
		\centering
		\includegraphics[width=5.9cm]{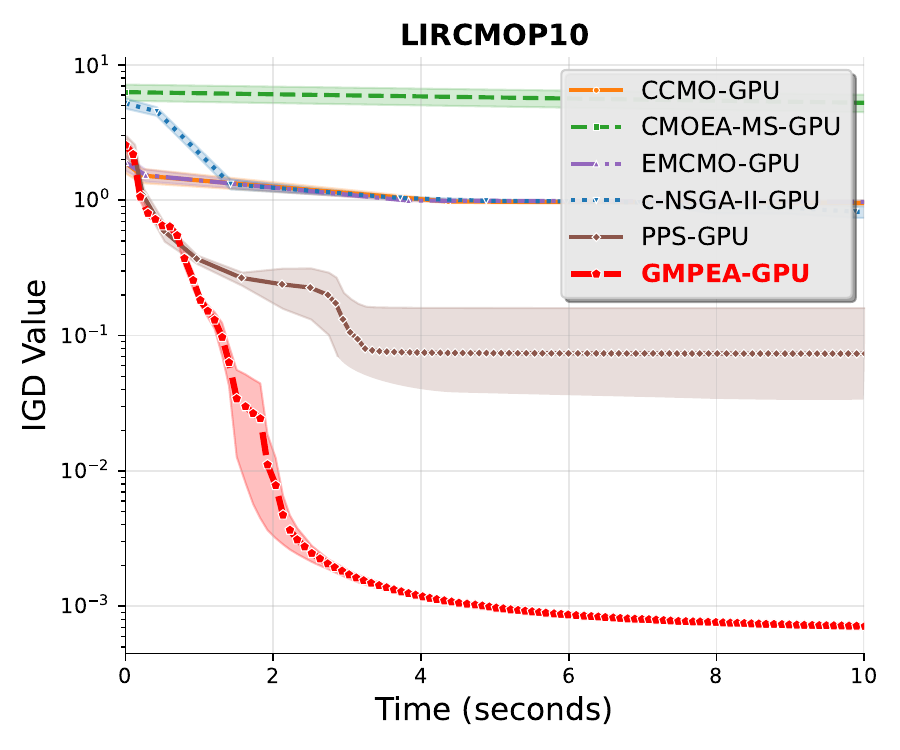}
		\text{(b)}
	\end{minipage}
	\begin{minipage}{5.8cm}
		\centering
		\includegraphics[width=5.9cm]{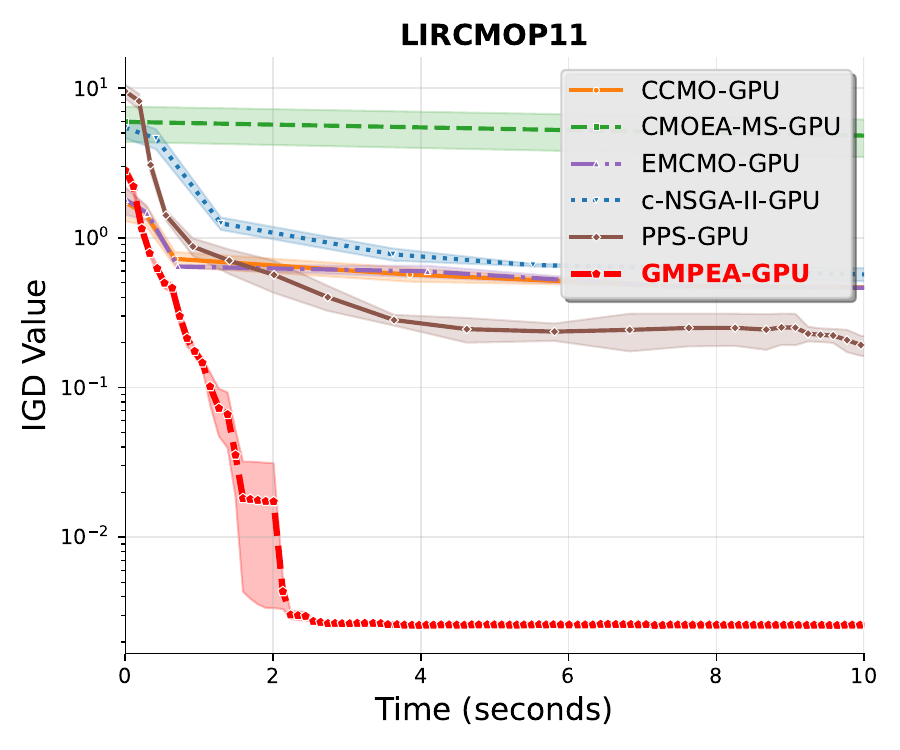}
		\text{(c)}
	\end{minipage}
	\centering
	
	\caption{Average IGD convergence curves for all algorithms on problems (a) LIRCMOP9, (b) LIRCMOP10, and (c) LIRCMOP11 under a fixed solving time. Shaded areas represent the upper and lower standard deviation ranges.}
	\label{f6}
\end{figure*}

Fig. \ref{f5} further illustrates the populations for the LIRCMOP9 and LIRCMOP11 problem under the 10-second fixed-time experiment. The figure shows that the solution set obtained by GMPEA not only converges to the true Pareto front but is also uniformly distributed along it, demonstrating excellent convergence and diversity. In comparison, the solution sets of other algorithms are sparse or fail to fully converge, with some algorithms barely completing their initialization phase. This visual evidence further emphasizes GMPEA's remarkable capability in handling time-sensitive CMOPs. The IGD convergence curves in Fig. \ref{f6} further reinforce this contrast.

\begin{table*}[htbp]
	\centering
	\setlength{\tabcolsep}{3pt}
	\caption{Comparison of Average IGD and Standard Deviation for Different Algorithms on C-DTLZ, DC-DTLZ, and LIRCMOP Test Suites under a Fixed Solving Time.}
    \begin{tabular}{p{5.28em}|c|c|c|c|c|c}
	\toprule
	\multicolumn{1}{c|}{\textbf{Problem}} & \textbf{C-NSGA2-GPU} & \textbf{PPS-GPU } & \textbf{CMOEA-MS-GPU } & \textbf{CCMO-GPU} & \textbf{EMCMO} & \textbf{GMPEA-GPU} \\
	\midrule
	\multicolumn{1}{c|}{C1-DTLZ1} & 5.26584 ± 1.57277 - & 0.01227 ± 0.00032 - & 13.56350 ± 2.39073 - & 0.00794 ± 0.00062 - & 0.00749 ± 0.00037 - & \textbf{0.00573 ± 0.00003 } \\
	\multicolumn{1}{c|}{C1-DTLZ3} & 0.12629 ± 0.12009 - & 0.05178 ± 0.00042 - & 23.17663 ± 3.25791 - & 0.01928 ± 0.00042 - & 0.01812 ± 0.00024 - & \textbf{0.01517 ± 0.00001 } \\
	\multicolumn{1}{c|}{C2-DTLZ2} & 0.01642 ± 0.00022 - & 0.02348 ± 0.00037 - & 0.13596 ± 0.00518 - & 0.01523 ± 0.00148 - & 0.01421 ± 0.00033 - & \textbf{0.01366 ± 0.00001 } \\
	\multicolumn{1}{c|}{C3-DTLZ4} & 0.07012 ± 0.00366 - & 0.06180 ± 0.05285 - & 0.37968 ± 0.04378 - & 0.06646 ± 0.00446 - & 0.06203 ± 0.00229 - & \textbf{0.02510 ± 0.00009 } \\
	\multicolumn{1}{c|}{DC1-DTLZ1} & 0.00416 ± 0.00005 - & 0.00956 ± 0.00313 - & 64.88421 ± 7.22499 - & 0.00562 ± 0.00035 - & 0.00553 ± 0.00040 - & \textbf{0.00535 ± 0.00001 } \\
	\multicolumn{1}{c|}{DC1-DTLZ3} & 0.01576 ± 0.00030 - & 0.02909 ± 0.00043 - & 22.10751 ± 3.93045 - & 0.01511 ± 0.00023 - & 0.01556 ± 0.00020 - & \textbf{0.01454 ± 0.00001 } \\
	\multicolumn{1}{c|}{DC2-DTLZ1} & 149.07948 ± 25.23735 - & 0.0119 ± 0.00171 - & 15.72792 ± 3.10152 - & 0.00814 ± 0.00105 - & 0.00743 ± 0.00024 - & \textbf{0.00581 ± 0.00003 } \\
	\multicolumn{1}{c|}{DC2-DTLZ3} & 88.38750 ± 26.05607 - & 0.04340 ± 0.00086 - & 19.18772 ± 2.52313 - & 0.01900 ± 0.00075 - & 0.01885 ± 0.00084 - & \textbf{0.01515 ± 0.00002 } \\
	\multicolumn{1}{c|}{DC3-DTLZ1} & 7.82105 ± 4.47117 - & 0.00819 ± 0.00188 - & 14.17995 ± 1.07805 - & 0.00649 ± 0.00030 - & 0.00640 ± 0.00021 - & \textbf{0.00465 ± 0.00001 } \\
	\multicolumn{1}{c|}{DC3-DTLZ3} & 109.88210 ± 3.05049 - & 0.02833 ± 0.00231 - & 209.27884 ± 29.17295 - & 0.00741 ± 0.00068 - & 0.00720 ± 0.00025 - & \textbf{0.01353 ± 0.00001 } \\
	\midrule
	\multicolumn{1}{c|}{LIRCMOP1} & 0.02913 ± 0.00161 - & 0.63232 ± 0.03004 - & 0.44269 ± 0.02926 - & 0.02700 ± 0.00169 - & 0.01996 ± 0.00145 - & \textbf{0.00294 ± 0.00071 } \\
	\multicolumn{1}{c|}{LIRCMOP2} & 0.02498 ± 0.00096 - & 0.21959 ± 0.26586 - & 0.22776 ± 0.05124 - & 0.01096 ± 0.00092 - & 0.00996 ± 0.00035 - & \textbf{0.00209 ± 0.00008 } \\
	\multicolumn{1}{c|}{LIRCMOP3} & 0.03914 ± 0.00563 - & 0.68205 ± 0.00615 - & 0.03812 ± 0.00284 - & 0.01295 ± 0.00136 - & 0.00951 ± 0.00036 - & \textbf{0.00315 ± 0.00116 } \\
	\multicolumn{1}{c|}{LIRCMOP4} & 0.03859 ± 0.00766 - & 0.66754 ± 0.00320 - & 0.04539 ± 0.00672 - & 0.01227 ± 0.00116 - & 0.00975 ± 0.00097 - & \textbf{0.00470 ± 0.00188 } \\
	\multicolumn{1}{c|}{LIRCMOP5} & 1.01101 ± 0.26081 - & 0.00935 ± 0.00801 - & 10.43762 ± 1.13898 - & 0.39382 ± 0.07142 - & 0.33256 ± 0.05369 - & \textbf{0.00113 ± 0.00004 } \\
	\multicolumn{1}{c|}{LIRCMOP6} & 0.62685 ± 0.36109 - & 0.00456 ± 0.00127 - & 10.82663 ± 0.95157 - & 0.41689 ± 0.00713 - & 0.44315 ± 0.01877 - & \textbf{0.00117 ± 0.00006 } \\
	\multicolumn{1}{c|}{LIRCMOP7} & 0.86592 ± 0.66870 - & 0.09767 ± 0.00987 - & 8.72703 ± 1.46787 - & 0.02187 ± 0.01376 - & 0.01628 ± 0.01191 - & \textbf{0.03883 ± 0.04080 } \\
	\multicolumn{1}{c|}{LIRCMOP8} & 0.58580 ± 0.21319 - & 0.08509 ± 0.03416 - & 11.39664 ± 1.21255 - & 0.10556 ± 0.17513 - & 0.20251 ± 0.18121 - & \textbf{0.03290 ± 0.03996 } \\
	\multicolumn{1}{c|}{LIRCMOP9} & 0.60909 ± 0.01579 - & 0.76742 ± 0.00373 - & 6.22916 ± 0.91429 - & 0.61585 ± 0.00582 - & 0.61967 ± 0.00900 - & \textbf{0.00075 ± 0.00005 } \\
	\multicolumn{1}{c|}{LIRCMOP10} & 0.86533 ± 0.08003 - & 0.09830 ± 0.07964 - & 4.95950 ± 0.31916 - & 0.97615 ± 0.02987 - & 0.98615 ± 0.01901 - & \textbf{0.00074 ± 0.00001 } \\
	\multicolumn{1}{c|}{LIRCMOP11} & 0.60062 ± 0.05926 - & 0.09476 ± 0.06763 - & 5.79684 ± 0.52482 - & 0.46122 ± 0.00830 - & 0.46219 ± 0.00426 - & \textbf{0.00256 ± 0.00004 } \\
	\multicolumn{1}{c|}{LIRCMOP12} & 0.48362 ± 0.10063 - & 0.00894 ± 0.00378 - & 5.84468 ± 0.51108 - & 0.21959 ± 0.06258 - & 0.25367 ± 0.02458 - & \textbf{0.00275 ± 0.00004 } \\
	\multicolumn{1}{c|}{LIRCMOP13} & 1.10077 ± 0.28363 - & 0.06593 ± 0.00463 - & 8.10710 ± 0.69971 - & 9.11583 ± 1.73369 - & 8.71970 ± 1.99504 - & \textbf{0.03508 ± 0.00028 } \\
	\multicolumn{1}{c|}{LIRCMOP14} & 0.83621 ± 0.31291 - & 0.07811 ± 0.00313 - & 9.03329 ± 0.95070 - & 8.93912 ± 1.03270 - & 7.97552 ± 1.60362 - & \textbf{0.02728 ± 0.00008 } \\
	\midrule
	Wilcoxon-Test(+/-/=) & 0/24/0 & 0/24/0 & 0/24/0 & 0/24/0 & 0/24/0 &  \\
	\bottomrule
\end{tabular}%
	\label{t4}%
\end{table*}%

\subsection{Results on Weapon-Target Assignment Problems}
The Weapon-Target Assignment (WTA) problem \cite{kline2019weapon} is a quintessential example of a time-sensitive military problem, typically requiring a solution within just a few seconds. To validate our approach, we adapted a two-objective and multi-constrained model from the work of \cite{li2024knowledge}. In this paper, we model the WTA problem as CMOPs. The first objective, $f_1$, minimizes the probability of surviving targets, which is equivalent to maximizing the overall target interception probability. The second objective, $f_2$, minimizes the consumption of weapon resources. The problem is formulated as follows:
\begin{equation}
	\begin{aligned}
		\mathrm{Minimize} \quad & \begin{cases} f_1(\mathbf{x}) = \sum_{i\in\mathbf{I}}\left(1-\prod_{k=1}^{N_K^i}\left(1-p_i^k\sum_{m\in\mathbf{M}}x_{im}^k\right)\right) \\ f_2(\mathbf{x}) = \sum_{m\in\mathbf{M}}x_{im}^k \end{cases} \\
		\mathrm{s.t.} \quad & \mathbf{x} = (x_{im}^k, y_{ir}^k, t_i^k) \in \Omega, \quad \forall i,m,r,k
	\end{aligned}
\end{equation}
The decision variables are denoted by $\mathbf{x}=(x_{im}^k, y_{ir}^k, t_i^k)$. Here, $\mathbf{I}$ is the set of $N_I$ enemy targets, and $\mathbf{M}$ is the set of $N_M$ deployed land-based missile vehicles. A value of $x_{im}^k = 1$ indicates that target $i$ is assigned to missile vehicle $m$ for the $k$-th strike, otherwise $x_{im}^k = 0$. Similarly, $y_{ir}^k = 1$ means target $i$ is assigned to radar $r$ for the $k$-th strike. The variable $t_i^k$ defines the strike time for target $i$, and $p_i^{k}$ is the probability of target $i$ being intercepted by its $k$-th assigned missile. In addition to the objective functions, the solutions must also satisfy various radar and launcher-related constraints detailed in \cite{li2024knowledge}. Based on the problem parameters from the original paper, the model is divided into 10 distinct scenarios, P1 through P10, for which we set a solving time limit of 10 seconds for our experiments.

\begin{figure}[!htbp]
	\centering
	\includegraphics[width=8.8cm]{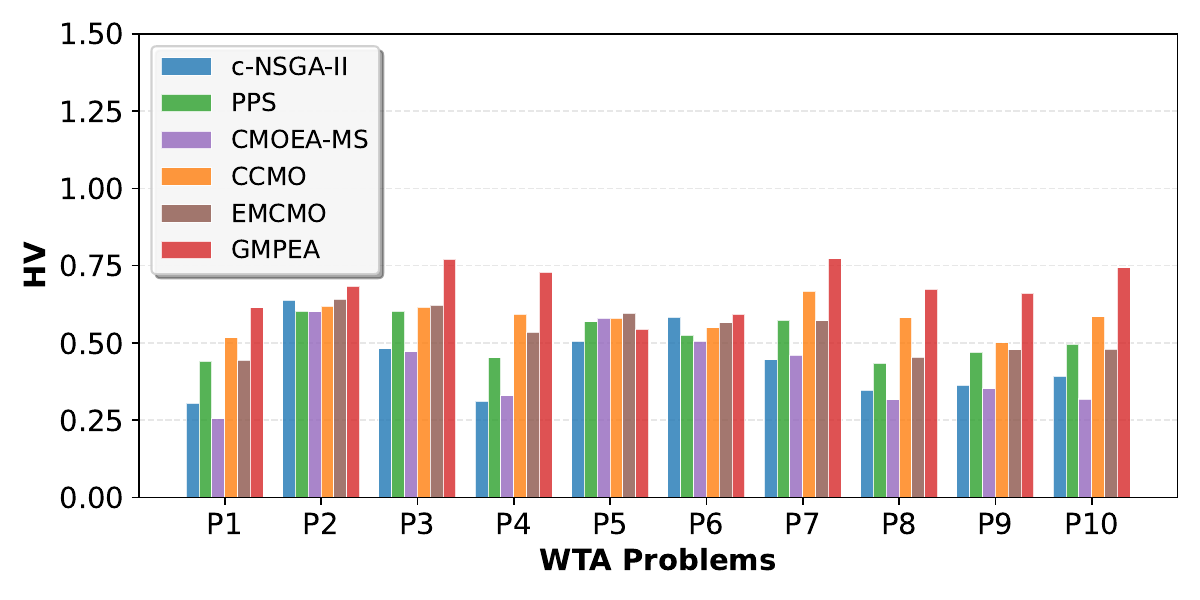}
	\centering
	
	\caption{Comparison of Average HV Results for All Algorithms on the Weapon-Target Assignment  Problems.}
	\label{f8}
\end{figure}

Fig. \ref{f8} shows a comparison of the Hypervolume (HV) results for all algorithms across these 10 problems. As presented in Table \ref{t5}, GMPEA achieved the best average HV results and standard deviations on 9 out of the 10 problems. Notably, for specific scenarios like P3 and P4, our proposed method significantly outperformed all other comparison algorithms. This outstanding performance within a short time frame highlights the high efficiency and strong problem-solving capability of GMPEA. Given the time-sensitive nature of this classic CMOP, we specifically focused on the algorithms' performance under a fixed time budget. The experimental results confirm that within a 10-second solving time, GMPEA consistently finds high-quality solution sets, with its HV metric being significantly superior to other algorithms on most problems. GMPEA's rapid solving capability is critically important for real-time military applications, further demonstrating that our method is not only effective on benchmark problems but is also highly competitive in complex, real-world applications.

\begin{table*}[htbp]
	\centering
	\caption{Comparison of Average HV result and Standard Deviation for Different Algorithms on WTA Problems under a Fixed Solving Time.}
	\begin{tabular}{ccccccc}
				\midrule
		& \textbf{c-NSGA-II-GPU} & \textbf{PPS-GPU} & \textbf{CMOEA-MS-GPU} & \textbf{CCMO-GPU} & \textbf{EMCMO-GPU} & \textbf{GMPEA-GPU} \\
		\midrule
		\textbf{P1} & 0.3047 ± 0.0300 & 0.4416 ± 0.0158 & 0.2548 ± 0.0291 & 0.5170 ± 0.0136 & 0.4435 ± 0.0161 & \textbf{0.6133 ± 0.0115} \\
		\textbf{P2} & 0.6375 ± 0.0088 & 0.6034 ± 0.0107 & 0.6003 ± 0.0078 & 0.6184 ± 0.0086 & 0.6408 ± 0.0103 & \textbf{0.6842 ± 0.0058} \\
		\textbf{P3} & 0.4820 ± 0.0136 & 0.6020 ± 0.0115 & 0.4720 ± 0.0112 & 0.6156 ± 0.0065 & 0.6215 ± 0.0100 & \textbf{0.7703 ± 0.0030} \\
		\textbf{P4} & 0.3122 ± 0.0190 & 0.4515 ± 0.0219 & 0.3296 ± 0.0202 & 0.5932 ± 0.0119 & 0.5340 ± 0.0111 & \textbf{0.7291 ± 0.0057} \\
		\textbf{P5} & 0.5048 ± 0.0104 & 0.5681 ± 0.0096 & 0.5799 ± 0.0292 & 0.5800 ± 0.0118 & \textbf{0.5965 ± 0.0121} & 0.5444 ± 0.0152 \\
		\textbf{P6} & 0.5831 ± 0.0121 & 0.5247 ± 0.0112 & 0.5050 ± 0.0157 & 0.5487 ± 0.0095 & 0.5650 ± 0.0130 & \textbf{0.5921 ± 0.0187} \\
		\textbf{P7} & 0.4464 ± 0.0148 & 0.5728 ± 0.0121 & 0.4602 ± 0.0117 & 0.6671 ± 0.0076 & 0.5721 ± 0.0093 & \textbf{0.7719 ± 0.0034} \\
		\textbf{P8} & 0.3471 ± 0.0225 & 0.4352 ± 0.0163 & 0.3157 ± 0.0230 & 0.5819 ± 0.0080 & 0.4540 ± 0.0143 & \textbf{0.6735 ± 0.0049} \\
		\textbf{P9} & 0.3633 ± 0.0177 & 0.4692 ± 0.0113 & 0.3526 ± 0.0204 & 0.5014 ± 0.0127 & 0.4781 ± 0.0299 & \textbf{0.6609 ± 0.0056} \\
		\textbf{P10} & 0.3926 ± 0.0198 & 0.4960 ± 0.0163 & 0.3188 ± 0.0188 & 0.5848 ± 0.0096 & 0.4794 ± 0.0142 & \textbf{0.7440 ± 0.0040} \\
		\bottomrule
	\end{tabular}%
	\label{t5}%
\end{table*}%

\subsection{Additional Experiments}

\subsubsection{GPU vs. CPU Time Analysis}
\begin{figure}[!htbp]
	\centering
	\includegraphics[width=8.8cm]{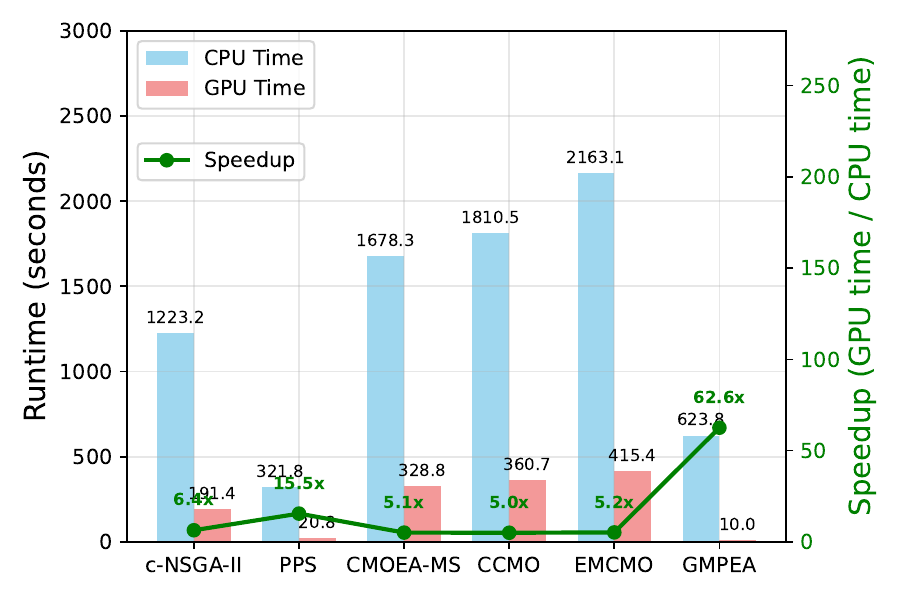}
	\centering
	
	\caption{Comparison of the average runtime of all algorithms on CPU and GPU.}
	\label{f7}
\end{figure}

Fig. \ref{f7} compares GMPEA and its competitors, showing their average runtimes on both CPU and GPU as well as the resulting speedups across all test problems.

The results clearly demonstrate a substantial performance gap. Traditional algorithms such as c-NSGA-II, PPS, CMOEA-MS, CCMO, and EMCMO exhibit relatively minor speedups when ported to the GPU, ranging from 5.0x to 15.5x. Although PPS shows the highest speedup among them, its overall runtime remains far higher than GMPEA's. This is because these algorithms, while running on a GPU, still rely on CPU-centric sequential paradigms for key operations like non-dominated sorting and environmental selection. These serial bottlenecks prevent them from fully utilizing the massive parallel processing power of the GPU. In stark contrast, GMPEA achieves a remarkable speedup of 62.6x over its CPU counterpart, which takes 623.8 seconds. This exceptional performance is a direct result of GMPEA's fully tensorized design, which redefines all core operations as highly parallelizable tensor computations. By fundamentally avoiding serial bottlenecks and converting conditional logic into element-wise masks, GMPEA's design is inherently suited for GPU acceleration. This allows it to efficiently utilize thousands of CUDA cores, drastically reducing runtime and significantly outperforming existing methods.

\subsubsection{Ablation Study}
To delve deeper into the key components of the GMPEA design, we conducted an ablation study specifically analyzing the impact of its differentiated neighborhood strategy on algorithm performance. In this experiment, we tested two variants: one where both populations used a large neighborhood (GMPEA-L) and another where both used a small neighborhood (GMPEA-S), comparing them against the original GMPEA ($Pop_1$ with a small neighborhood, $Pop_2$ with a large one). As shown in Table \ref{t6} , the original GMPEA achieved the best IGD values on most problems. The results indicate that while GMPEA-L performed well on some problems, its overall performance was slightly inferior to the original. This is primarily because a single large-neighborhood strategy, while promoting rapid convergence, sacrifices local exploration, leading to a decline in performance on problems with complex constraints or multi-modal characteristics. Conversely, GMPEA-S performed even worse on some problems. A single small-neighborhood strategy, while enhancing local diversity, has a slower convergence rate and is prone to getting trapped in local optima. These findings strongly confirm that GMPEA's differentiated neighborhood strategy is crucial to its success. By assigning a small neighborhood to $Pop_1$ to enhance local search for constraint handling, and a large neighborhood to $Pop_2$ to accelerate global convergence, GMPEA cleverly balances convergence and diversity, achieving optimal overall performance.

\begin{table}[htbp]
	\centering
	\setlength{\tabcolsep}{3pt}
	\caption{Statistical Results of Average IGD and Standard Deviation for the Proposed GMPEA and Its Two Variants on the LIRCMOP Test Suite.}
	\begin{tabular}{cccr}
		\toprule
		\textbf{Problem} & \textbf{GMPEA-S} & \textbf{GMPEA-L} & \multicolumn{1}{c}{\textbf{GMPEA}} \\
		\midrule
		LIRCMOP1 & 0.00332±0.00050 - & 0.00335±0.00056 - & \multicolumn{1}{c}{\textbf{0.00294±0.00071 }} \\
		LIRCMOP2 & 0.00299±0.00022 - & 0.00288±0.00049 - & \multicolumn{1}{c}{\textbf{0.00209±0.00008 }} \\
		LIRCMOP3 & 0.01174±0.00965  & 0.00815±0.00325 - & \multicolumn{1}{c}{\textbf{0.00315±0.00116 }} \\
		LIRCMOP4 & 0.00419±0.00415 + & \textbf{0.00419±0.00286 +} & \multicolumn{1}{c}{0.00470±0.00188 } \\
		LIRCMOP5 & 0.00386±0.00560 - & 0.00149±0.00009 = & \multicolumn{1}{c}{\textbf{0.00113±0.00004 }} \\
		LIRCMOP6 & 0.00177±0.0007 = & 0.00129±0.00003 = & \multicolumn{1}{c}{\textbf{0.00111±0.00003 }} \\
		LIRCMOP7 & 0.06508±0.03803 - & 0.05605±0.05111 - & \multicolumn{1}{c}{\textbf{0.03643±0.03701 }} \\
		LIRCMOP8 & 0.05393±0.02676 - & 0.03922±0.04878 - & \multicolumn{1}{c}{\textbf{0.03290±0.03996 }} \\
		LIRCMOP9 & 0.00079±0.00002 = & 0.00119±0.00005 = & \multicolumn{1}{c}{\textbf{0.00075±0.00005 }} \\
		LIRCMOP10 & 0.00093±0.00001 = & 0.00121±0.00004 - & \multicolumn{1}{c}{\textbf{0.00072±0.00001 }} \\
		LIRCMOP11 & 0.00261±0.00004 = & 0.00260±0.00005 = & \multicolumn{1}{c}{\textbf{0.00254±0.00004 }} \\
		LIRCMOP12 & 0.00320±0.00007 - & 0.00320±0.00010 - & \multicolumn{1}{c}{\textbf{0.00275±0.00004 }} \\
		LIRCMOP13 & 0.04713±0.00080 - & 0.05948±0.00136 - & \multicolumn{1}{c}{\textbf{0.03361±0.00020 }} \\
		LIRCMOP14 & 0.02782±0.00014 = & 0.03166±0.00040 - & \multicolumn{1}{c}{\textbf{0.02700±0.00004 }} \\
		\midrule
		Wilcoxon-\\Test(+/-/=) & 1/8/5 & 1/9/4 &  \\
		\bottomrule
	\end{tabular}%
	\label{t6}%
\end{table}%

\subsubsection{Time Consumption with Different Population Sizes}

\begin{figure}[!htbp]
	\centering
	\includegraphics[width=8.8cm]{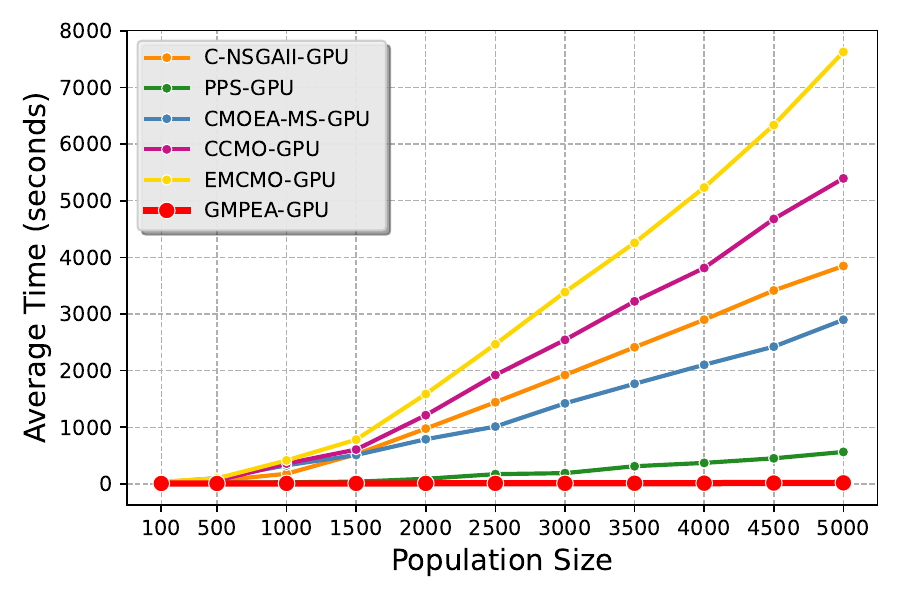}
	\centering
	
	\caption{Average runtime trends of all algorithms with varying population sizes.}
	\label{f9}
\end{figure}
To analyze the algorithms' sensitivity to population size, we conducted an additional set of experiments where the population size was varied as follows: 100,500,1000,1500,2000,2500,3000,3500,4500,5000.

Fig. \ref{f9} illustrates how the average runtime of GMPEA and its competitors changes with varying population sizes. The graph clearly shows that: GMPEA's time consumption increases with population size, but its growth rate is far lower than that of all other algorithms. Even with a population size as large as 5000, GMPEA's runtime remains low. This demonstrates that GMPEA's fully tensorized design makes its computational complexity highly insensitive to population size, enabling it to efficiently handle large-scale populations.

The time consumption of other algorithms shows a distinct non-linear increase as the population size grows. Algorithms relying on serial strategies like non-dominated sorting or truncation exhibit a steep rise in runtime. This further validates our central argument that the computational efficiency bottleneck of existing CMOEAs lies in the mismatch between their algorithm structure and the parallel architecture of GPUs. This result underscores GMPEA's unique advantage in handling large-scale populations, positioning it as a powerful tool for solving high-dimensional and large-scale CMOPs.

\section{Conclusion}
\label{s6}
This paper addresses the bottlenecks in computational speed and efficiency that hinder existing CMOEAs on time-sensitive problems. We propose GMPEA.  The core innovation of this algorithm is its fully tensorized architectural design, which deeply integrates multi-population co-evolution, a differentiated neighborhood strategy, and efficient parallel computing. GMPEA abandons the common serially dependent environmental selection strategies found in existing CMOEAs, instead adopting a new, decomposition-based tensorized environmental selection process. This achieves complete tensorization from initialization to environmental selection.

The superiority of GMPEA has been fully validated through extensive experiments on multiple CMOP benchmark suites and  real-world weapon-target assignment problems. Experimental results show that under the same GPU acceleration, GMPEA's solution performance is comparable to current mainstream CMOEAs, even without a time limit, but its solving speed is tens to hundreds of times faster. More importantly, in time-sensitive scenarios with a strict 10-second solving time limit, GMPEA's performance far surpasses all competitors. This powerfully demonstrates its unique advantage in solving time-sensitive CMOPs. Furthermore, ablation studies confirm that GMPEA's differentiated neighborhood strategy is key to its superior performance, while time consumption analysis reveals its low sensitivity to large population sizes, showcasing its excellent scalability.

While GMPEA has made significant progress in solving time-sensitive CMOPs, there are still some directions worth exploring in the future. Future research can explore how to extend GMPEA to effectively handle dynamic and uncertain CMOPs. This may involve introducing new population adaptation mechanisms or predictive models to quickly respond to environmental changes or handle parameter uncertainties. Simultaneously, as the number of objectives in CMOPs increases, solving many-objective CMOPs becomes more challenging. Although GMPEA's decomposition-based framework offers a potential advantage for solving many-objective problems, a deeper study is needed on balancing convergence and diversity when dealing with high-dimensional objective spaces. Future work will focus on designing more efficient tensorized environmental selection strategies for many-objective CMOEAs and exploring new diversity maintenance mechanisms to fully leverage the parallel capabilities of GPUs.

\ifCLASSOPTIONcaptionsoff
  \newpage
\fi

\bibliography{reference}

\end{document}